\documentclass[11pt,a4paper]{article}
\usepackage[T1]{fontenc}
\usepackage[utf8]{inputenc}
\usepackage{lmodern}
\usepackage[margin=2.5cm]{geometry}
\usepackage{microtype}
\usepackage{booktabs}
\usepackage{graphicx}
\usepackage{url}
\usepackage[numbers,sort&compress]{natbib}
\usepackage[hidelinks]{hyperref}
\usepackage{caption}
\newcommand{\mname}[1]{#1}
\newcommand{\orcid}[1]{\href{https://orcid.org/#1}{#1}}
\newcommand{\nonum}[1]{#1}
\DeclareUnicodeCharacter{2032}{\ensuremath{'}}

\title{norregel: a preregistered benchmark of dated numeric values in Norwegian law}
\author{Eirik Botten Nicolaysen\\EcoDeco AS\\ORCID \orcid{0009-0001-9188-6788}\\eirik@ecodeco.no}
\date{}

\begin{document}
\maketitle

\begin{abstract}
norregel asks language models for numeric values in Norwegian statutes, regulations and parliamentary decisions:
NOK amounts, percentages, multiples of the National Insurance basic amount (G) and age limits, each at a stated
date. The 916 items and their keys were extracted mechanically from Lovdata's open data. Each item carries the dated
history of its provision, so an answer can be classed as correct, outdated, wrong or uninterpretable. This paper
reports a first and a second preregistered run on the same items. In the first run, the list of five models was
published only after the runs had been analysed locally. H1, that outdated answers are more frequent for values that
took effect after a model's cutoff, could not be tested in substance, and the tests of H2 and H3 were inconclusive.
\mname{Opus 5.5} and \mname{Sonnet 4.6} answered 46.7\% and 40.2\% of the 880 primary items correctly, and most wrong
answers were values not in the item's dated history. The second run was preregistered and published before any of its
calls. Its models were chosen by a criterion stated in advance, and eight models were analysed. Its pooled primary
test of H1 rejected the null hypothesis ($d = 17.8$ percentage points, 95\% interval [7.7, 28.9], $p = 0.0001$), so H1
is supported under the preregistered decision rule. The preregistered sensitivity analysis S8 counts as outdated only
the most recent value that differs from the key, which gives both groups one target value. In S8, $d$ was 3.9
[$-$3.6, 12.7] ($p = 0.1754$), an interval that contains 0. H1 was supported in every other preregistered sensitivity analysis, but S8 is the only one that gives both groups one
target value, and under it the effect falls from 17.8 to 3.9 percentage points and is not established. We therefore do
not regard the support for H1 as robust. H2 was reported as an estimate: the share of outdated answers
equal to the most recent value that differs from the key was $s = 56.1\%$ [46.6, 65.0], an interval that contains
0.5. The pooled H3 test, which asked whether the share correct is lower in Nynorsk than in Bokmål, was inconclusive
($D = -0.4$ percentage points [$-$1.1, 0.3]). In both runs some models gave mostly uninterpretable answers.
\mname{Haiku 4.5} is among them: uninterpretable answers made up 94.0\% of its 880 answers on the primary set of the
main task in the first run, 93.1\% of its 880 answers on the primary set of the main task in the second run, and
93.4\% of its 1\,760 answers in the H3 task of the second run.
\end{abstract}

\section{Introduction}

Many Norwegian rules are numbers that change over time. The basic amount G and the student-support rates form yearly
series, and tax rates are set in the annual tax decision. Asked for such a value at a given date, a language model can
give the value in force at that date, a value that applied earlier, or a value that never applied. Which of these it
gives is the first research question of the preregistration. Language models are trained on data collected up to a
certain point. Facts with an expiry date are a known problem for them \citep{dhingra2021timeaware}, and language
models have been shown to perform worse on text from beyond their training period \citep{lazaridou2021mindthegap}.
Hallucinated legal facts have been documented for US federal case law \citep{dahl2024fictions}.

norregel measures this for Norwegian law. Each item asks for a single value at a stated date, and the scorer compares
the answer with the dated history of the item's provision. This is what separates an outdated answer from a wrong \nonum{one}.
The paper describes the benchmark (Section~\ref{sec:build}), the preregistration and run protocol of the first run
(Section~\ref{sec:prereg}), the scorer (Section~\ref{sec:score}), the first run (Section~\ref{sec:runs}) and its
results for five models (Section~\ref{sec:results}). It then describes the second run, preregistered after the first so
that H1 could be tested (Section~\ref{sec:v2}). Section~\ref{sec:dev} lists all deviations from both preregistrations,
and Section~\ref{sec:discussion} states what the results show and what they do not show.

\section{Related work}

\paragraph{Norwegian and Nordic benchmarks.} NorBench \citep{samuel2023norbench} and ScandEval
\citep{nielsen2023scandeval} benchmark Norwegian and Scandinavian language models. NorEval \citep{mikhailov2025noreval}
covers both written standards of Norwegian, Bokmål and Nynorsk, and NLEBench \citep{liu2023nlebench} evaluates
Norwegian text generation. norregel is narrower: every item asks for a value that a legal source fixes at a stated
date, and the key comes from that source.

\paragraph{Time-dependent questions.} TimeQA \citep{chen2021timeqa} builds time-sensitive questions from facts in
WikiData that change over time. SituatedQA \citep{zhang2021situatedqa} studies questions whose answers depend on when
and where they are asked. StreamingQA \citep{liska2022streamingqa} asks questions on a given date and answers them from
time-stamped news. RealTime QA \citep{kasai2022realtimeqa} announces questions about current events and evaluates
systems at regular intervals; it reports that GPT-3 tends to return outdated answers when the retrieved documents are
insufficient. FreshQA \citep{vu2023freshllms} is a dynamic benchmark that includes questions that require
fast-changing world knowledge. TemporalWiki \citep{jang2022temporalwiki} uses the difference between consecutive
snapshots of Wikipedia and Wikidata to track whether a model retains old knowledge and acquires new knowledge.
\citet{dhingra2021timeaware} probe language models for facts that change over time, and \citet{lazaridou2021mindthegap}
show that Transformer-XL language models perform worse on utterances from beyond their training period.

\paragraph{Legal evaluation.} LegalBench \citep{guha2023legalbench} is a collaboratively built benchmark for legal
reasoning in large language models. \citet{magesh2024hallucination} report a preregistered evaluation of AI legal
research tools and find that the providers' claims about hallucinations are overstated.

\paragraph{Knowledge cutoffs and refusals.} H1 relies on the training-data cutoff stated by the provider.
\citet{cheng2024dated} report that ``effective cutoffs often differ from reported cutoffs''. \citet{kadavath2022know}
study whether language models can predict which questions they will answer correctly, and \citet{yin2023selfaware}
test whether they can identify unanswerable questions. norregel scores a refusal as uninterpretable.

\paragraph{Preregistration.} Preregistration fixes the research questions and the analysis plan before the outcomes
are observed \citep{nosek2018prereg}. It has been proposed for NLP research \citep{vanmiltenburg2021prereg} and adapted
to predictive modelling \citep{hofman2023prereg}. The runs used Inspect \citep{inspect}.

\paragraph{Auditing frameworks.} SimpleAudit describes itself as ``a simple, extensible, local-first framework for
multilingual auditing and red-teaming of AI systems via adversarial probing'' \citep{simpleauditrepo}, and its
methodology paper is \citet{gautam2026benchmarkless}. norregel complements it with a closed set of dated facts.

\paragraph{A related tool by the author.} The author's pull request \#105 to SimpleAudit \citep{simpleaudit105},
opened on 9 October 2026 and merged on 10 October 2026, added a deterministic check of answers against dated facts.
The preregistration requires it to be named as earlier work. Candidate items that overlap its dated facts are kept in
norregel and marked (93 items in 6 series).

\section{Benchmark construction}\label{sec:build}

\paragraph{Source and extraction.} Candidates are extracted mechanically from Lovdata's open data (NLOD 2.0, \url{https://api.lovdata.no/v1/publicData}). Inneholder data under Norsk lisens for offentlige data (NLOD) tilgjengeliggjort av Stiftelsen Lovdata (contains data under the Norwegian Licence for Open Government Data, made available by Stiftelsen Lovdata). The core
documents were fixed before counting. The long tail is counted but not built as items in v1.

\paragraph{Year items.} The unit is the year item: one question per provision and year of validity. A series groups the
year items of a provision. Measures are computed on year items, and intervals use the series as cluster: a
nonparametric bootstrap that resamples whole series \citep{field2007bootstrapping}.

\paragraph{Cleanliness rule.} Only clean candidates are included (rule v3 with its amendments up to v3.6). In a flagged
series, a year is accepted only when the value and the rate's name or anchor phrase stand in the same clause of that
year's verbatim text; otherwise the series is split there. The last value must equal the current consolidated text.
Qualifiers in the same clause as the value, such as «statens andel» or «i tiltakssonen», belong to the identity of the
series. Valid-from and valid-to dates follow a fixed hierarchy: an explicit period or date in the text, then «med
virkning fra», then the date of entry into force. A year without a sourced date is discarded and splits the series.

\paragraph{Value types and set types.} v1 contains NOK amounts, percentages, G multiples and ages. Deadlines were
not included in either run (\path{docs/PREREG_V2.md:32}), so 21 clean deadline items in the core are not included. The annual tax decision and the decision on
social-security contributions are chained as named rates from a list fixed before extraction, and 4 duplicate chains
with 33 year items were removed. Table~\ref{tab:sets} gives the items per set type, which the preregistration promised
to report. In the primary set, the years of validity run from 2001 to 2026, and a series has between 1 and 26 items
(median 6.0).

\begin{table}[t]
\centering
\footnotesize
\caption{Items per set type (the field \texttt{sett}), all items and primary set, and series in the primary set. Not preregistered as a table; computed after review (supplement A2).}
\label{tab:sets}
\begin{tabular}{llll}
\toprule
Set type & Items (all) & Items (primary) & Series (primary) \\
\midrule
permanent provisions & 265 & 235 & 48 \\
yearly regulations & 92 & 92 & 21 \\
basic amount G & 26 & 26 & 1 \\
named rates & 533 & 527 & 26 \\
all & 916 & 880 & 96 \\
\bottomrule
\end{tabular}
\end{table}

\paragraph{Questions.} Each item was written from a fixed template per type and written standard, without a model and
without manual curation. The Bokmål template for an amount is «Hvor mange kroner var \{navn\} etter \{kilde\} per
\{dato\}?», followed by the instruction «Svar med verdien alene på siste linje, slik: SVAR: <verdi>».

\paragraph{Checks and locks.} In each round of a blind check, 40 accepted year items were extracted again by a
language-model sub-agent directed by the author, without access to the reference values. Round 1 gave 39/40, and the error, in the author's extraction, was corrected. Rounds 2 to 5 gave 40/40. These rounds were run on the data before a correction
of the age items, which changed the type of age limits and thereby which series are clean. The set was then locked
locally. The first, second and third locks were reopened, before any model had been run, because their final checks
found date errors: at the first, 24 items asked about a date before the value applied; at the second, 3 items had a
wrong key because a rate period had not been used; at the third, valid-from and valid-to dates agreed in 39/40. The
fourth lock is final under point 3 of cleanliness rule v3.6, which was adopted when the third lock was reopened: errors found after
it are recorded as errata, and it is not reopened. Its final check agreed on 40/40 values, 38/40 valid-from dates and
31/40 valid-to dates. The 2 valid-from differences belong to erratum E1: partial dates of the form «\ldots{} trer i
kraft straks med virkning fra og med inntektsåret ÅÅÅÅ» were not used, and a mechanical search found 10 items, all
in two series of the Tax Act, with a wrong valid-from date. The errata list 4 of them with a wrong key for the
question date; the audit of the fourth lock, made by language-model sub-agents directed by the author, found a fifth,
\texttt{norregel-00909}. These 5 items are excluded as
X1 (below). The other E1 items, \texttt{norregel-00907, -00908, -00910, -00911 and -00916}, have a wrong valid-from
date but the right key; they are flagged, kept in the primary set and left out in S3. Erratum E2 records a difference
in the definition of a period provision in 9 of the 40 checked items; the dates in these items are the same as the
sub-agent's, and no key is affected. The fourth lock has 916 items in 97 series, and 454 of the 916 items (49.6\%) are in
Nynorsk.

\paragraph{Exclusions and analysis sets.} The addendum to the preregistration defines exclusion groups: X1, 5 items with
a wrong key for the question date; X2, 23 items whose question date, 14 February 2027, lies after the corpus retrieval
on 9 October 2026; X3, 6 ambiguous items; and X4, 2 items affected by amendments of the annual tax decision. The
primary set is the 916 items minus these 36, that is 880 items in 96 series. Seven flagged items stay in the primary set and are left out in sensitivity analysis S3: the other E1 items named above, and \texttt{norregel-00133 and -00149} from the annual tax decision, whose dates are wrong but whose key is correct on the date in the question. In 219 items of the primary set the preceding value equals the key,
and in 234 some earlier value does; in these items an outdated answer cannot be told from a correct answer. The H1 set
has 567 items (primary items that have a preceding value that differs from the key), and the H2 set has 570 items (primary items with at least two distinct earlier values that differ from the key; an item can be in the H2 set even when its preceding value equals the key).
The classes of the excluded items are in Table~\ref{tab:groups} (Appendix~\ref{app:prereg}).

\paragraph{H3 pairs.} For H3, 200 of the 916 items were drawn with a fixed seed, stratified on series; each of the 97 series has at least one item. Each was
asked in both Bokmål and Nynorsk, with the same template and key, as 400 questions in 200 pairs. The H3 run is
separate from the main run. Pairs whose source item is excluded are left out, which leaves 189 pairs.

\section{Preregistration and protocol of the first run}\label{sec:prereg}

The preregistration, in Norwegian, and an English addendum were locked and published on Zenodo as v1 on 16 October
2026 at 16:18:04 UTC (DOI \texttt{10.5281/zenodo.23422303}), with four files: the preregistration, the addendum, the
lock manifest and the exclusion list. The run protocol and the model list were committed locally at 16:30:44 UTC and
published only in v1.1. The first log with benchmark items starts at 16:30:54 UTC (a pilot attempt that failed locally,
A1), the first answered pilot request at 16:31:39 UTC, and the main runs at 16:45:40 UTC. The addendum resolved what an
audit of the fourth lock, by language-model sub-agents directed by the author, had found: an item with a wrong key, an
exclusion rule that existed only as a recommendation, hypotheses that could not be run without asking the author, and
answer formats the locked scorer could not read. It defines scorer v2, the analysis sets, the tests and the run
configuration, and it names a reference implementation of the analysis, which was used unchanged.

The run protocol was written and committed before any model was run. It settles points that the addendum left open,
including the cutoff rule, vague cutoffs, complete runs and restarts, the «år og måneder» (years and months) rule and token and
temperature limits, and it fixes a pilot of 20 items and a cost cap of 60 USD. Changes during and after the runs are amendments A1--A8, each dated and committed. The deviations from the addendum and the run protocol are D1--D13 (Section~\ref{sec:dev}). Version v1.1
of the preregistration (DOI \texttt{10.5281/zenodo.23432124}), published on 17 October 2026, contains the protocol,
the model list and the Ollama Modelfiles. The concept DOI for all versions is \texttt{10.5281/zenodo.23422302}.

\paragraph{Deviation D12.} The addendum requires the model list to be published as a new version of the
preregistration before any result. The list was published in v1.1 after the runs and the analysis had been completed
locally, but before any result was made public. The list was committed locally before the first run (\texttt{3c554cff}). Rows 5 and 4
were changed afterwards (A2, A3), and A3 came after an \mname{NB-Llama 8B} run with 109 samples had been discarded.
The list that was run and published therefore differs from the committed list in these two rows. The commit history
is published with the results (Zenodo v1.2, \texttt{10.5281/zenodo.23434782}).

\section{Scoring}\label{sec:score}

Scorer v2 is the primary scorer; the locked scorer v1 is used in sensitivity analysis S2. The scorer reads the answer
line, which is the last line where the label «svar» is followed by a colon, with at most markdown characters before
it. On that line it removes bracketed text and markdown, and it requires exactly one number. Digits are read with
Norwegian and English separators; Norwegian number words from 0 to 99 count when the line has no digits; words such as
«kr», «prosent» or «G» are ignored. For an age item, an answer in months is wrong.

\paragraph{Classes.} Two numbers are equal when they differ by less than \texttt{1e-6}. An answer is \emph{correct}
when it equals the key, and \emph{outdated} when it equals the value of an earlier version in the item's history; the
scorer reports the age in years. An answer is \emph{wrong} when a number was read but it is neither. Wrong is split into a later value and a value not in the history. The latter also covers values from before the start of the clean
series, so it is an upper bound for a value that never applied. An answer is
\emph{uninterpretable} when there is no answer line, no number, or more than one number. Refusals, empty outputs and
samples still in error after the retries are uninterpretable. All shares use all answers as the denominator.

v1 uses a tolerance of 0.005 and counts later values as outdated, so H2 and FS2 are not defined under v1. For an age
limit, «X år og Y måneder» is read as X + Y/12 years before scoring (deviation D3); no answer had this form.

\paragraph{Validation.} Before any model run, scorer v2 was tested on all 49 answer-format edge cases from the audit
and on 800 synthetic answers, all classified as expected, and dry runs that returned the key as the answer scored 916 of
916 main items and 400 of 400 H3 questions correct (addendum).

\section{First run: models and runs}\label{sec:runs}

\begin{table*}[t]
\centering
\footnotesize\setlength{\tabcolsep}{4pt}
\caption{Models. Cutoff is the provider's stated training-data cutoff; a local model whose card states no date has an unknown cutoff and is not tested in H1.}
\label{tab:models}
\begin{tabular}{lllll}
\toprule
Model & Inspect model id & Cutoff & Reasoning & Run \\
\midrule
\mname{Haiku 4.5} & {\scriptsize\path{anthropic/claude-haiku-4-5-20251001}} & 2025-07-31 & no & API \\
\mname{Sonnet 4.6} & {\scriptsize\path{anthropic/claude-sonnet-4-6}} & 2026-01-31 & no & API \\
\mname{Opus 5.5} & {\scriptsize\path{anthropic/claude-opus-5-5}} & 2026-06-30 & yes & API \\
\mname{NB-Llama 8B} & {\scriptsize\path{ollama/nb-llama-3.1-8b-instruct-cardtmpl:latest}} & unknown & no & local (Ollama) \\
\mname{NorMistral 7B} & {\scriptsize\path{ollama/normistral-7b-warm-instruct-cardtmpl:latest}} & unknown & no & local (Ollama) \\
\bottomrule
\end{tabular}
\end{table*}

Table~\ref{tab:models} lists the models. They were chosen by the author before the runs, not in the preregistration; the addendum leaves the choice of models to run time. \mname{Haiku 4.5} and \mname{Sonnet 4.6} were chosen although newer Haiku and Sonnet versions (5.5) existed \citep{anthropicmodels}, and
\mname{Opus 5.5} was the newest Opus model in the provider's model list as read by the author on 16 October 2026; the
list was not stored (\path{docs/RUN_PROTOCOL.md:61}). No criterion for the choice of models was stated in advance.
The cutoffs and the reasoning classification follow the providers' model pages \citep{haiku45page,sonnet46page,opus55page};
MODELLER.md in v1.1 gives the URL, a verbatim quote and the SHA-256 of each fetched page. The local models are \mname{NB-Llama 8B} \citep{nbllamacard} and the instruct variant of
\mname{NorMistral 7B} \citep{normistralcard}. Neither card states a cutoff, so neither model is tested in H1.

\paragraph{Parameters.} No system prompt; one run per model and task; temperature 0 and at most 1\,024 output tokens
for non-reasoning models. \mname{Opus 5.5} is a reasoning model and was run without a temperature setting, with at most
16\,384 tokens and with effort and thinking at the provider's defaults. Inspect's own retries apply, and a sample still
in error is re-run up to 3 times. The local models ran with one connection (D2), and \mname{NorMistral 7B} also with a
limit of 2 provider-level retries from the eval-retry onwards (A6, D10): the first 91 samples of its main run ran
without that limit, and its H3 run used it throughout. Software: \texttt{Python 3.12.13}, \texttt{inspect\_ai 0.3.245},
\texttt{numpy 2.5.4} and, for the local models, \texttt{Ollama 0.35.1}.

\paragraph{Chat templates.} \mname{NorMistral 7B} was run with the chat template of its model card, as a new Ollama
model on the same weights (A2); a check before the first sample confirmed that the main Ollama service rendered the prompt in the card's format (A2, A4). \mname{NB-Llama 8B} was run with the card's template with a fixed date and the
begin-of-text token (A3). The main Ollama service selected the template stored in the GGUF file and ignored the
Modelfile; that template used the current date and no begin-of-text token. This was the reason for running
\mname{NB-Llama 8B} on a separate Ollama instance, which uses the Modelfile's template.

\paragraph{Completion and other runs.} All five models were run complete on the main task and on H3 under the protocol's rule for
complete runs. The only final errors (samples still in error after the retries, scored uninterpretable) were 3 samples
of \mname{NorMistral 7B} on the main task. Answers that reached the token limit were scored as they stood: 297 on the
main task and 138 on H3 for \mname{NB-Llama 8B}, of which 292 and 133 repeated a line at least three times (a definition set after the language-model review), and 10 and
11 for \mname{NorMistral 7B}. A pilot on 20 items, which
does not count, checked the parameters, and the rule for a rejected temperature was not triggered. The API runs cost
26.80 USD including the pilot, under the cap of 60 USD. The addendum requires every run to be reported:
Table~\ref{tab:otherruns} (Appendix~\ref{app:prereg}) lists the pilots, the discarded and the stopped runs with their
classes, none of which count in the results.

\section{First run: results}\label{sec:results}

\begin{table*}[t]
\centering
\footnotesize
\caption{Primary set, scorer v2: share of all answers (\%) with cluster-bootstrap CI in brackets. Percentages are rounded to one decimal, half to even (Python string formatting), so a row need not add up exactly. Figure~\ref{fig:classes} shows the shares.}
\label{tab:primary}
\begin{tabular}{lllll}
\toprule
Model & Correct & Outdated & Wrong & Uninterpretable \\
\midrule
\mname{Haiku 4.5} & 0.5 [0.0, 1.1] & 0.1 [0.0, 0.4] & 5.5 [3.2, 7.7] & 94.0 [91.5, 96.4] \\
\mname{Sonnet 4.6} & 40.2 [29.3, 49.7] & 6.5 [3.7, 9.6] & 40.9 [32.4, 51.0] & 12.4 [9.0, 16.7] \\
\mname{Opus 5.5} & 46.7 [35.4, 56.2] & 6.5 [4.0, 9.3] & 46.4 [37.1, 57.4] & 0.5 [0.1, 1.0] \\
\mname{NB-Llama 8B} & 8.5 [3.4, 14.1] & 1.2 [0.3, 2.5] & 21.6 [14.9, 28.3] & 68.6 [60.0, 78.0] \\
\mname{NorMistral 7B} & 0.0 [0.0, 0.0] & 0.0 [0.0, 0.0] & 0.2 [0.0, 0.6] & 99.8 [99.4, 100.0] \\
\bottomrule
\end{tabular}
\end{table*}

\begin{figure}[t]
\centering
\includegraphics[width=0.85\linewidth]{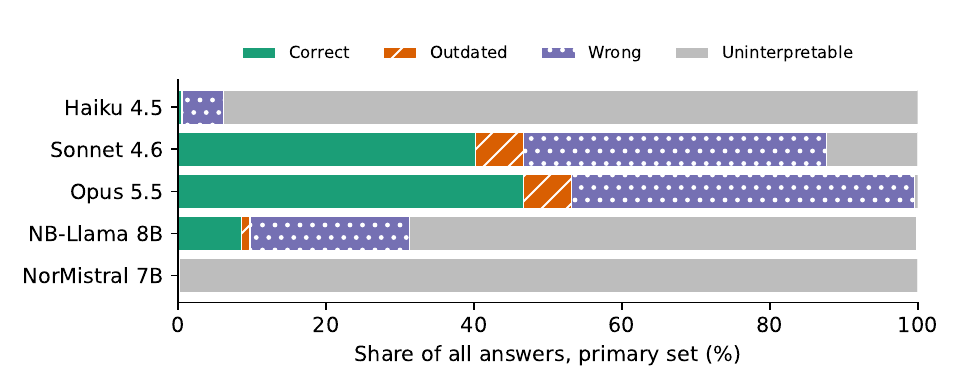}
\caption{Class shares per model on the primary set (scorer v2), from Table~\ref{tab:primary}.}
\label{fig:classes}
\end{figure}

\paragraph{Primary measures.} On the primary set (880 items, 96 series; Table~\ref{tab:primary},
Figure~\ref{fig:classes}), \mname{Opus 5.5} answered 46.7\% of the items correctly and \mname{Sonnet 4.6} 40.2\%;
\mname{NB-Llama 8B} 8.5\%, \mname{Haiku 4.5} 0.5\% and \mname{NorMistral 7B} 0.0\%. Outdated answers were 6.5\% for
\mname{Sonnet 4.6} and 6.5\% for \mname{Opus 5.5}, and at most 1.2\% for the other models. Of the 880 answers of each
model on the primary set, 94.0\% were uninterpretable for \mname{Haiku 4.5}, 68.6\% for \mname{NB-Llama 8B} and 99.8\%
for \mname{NorMistral 7B}. These shares are
reported as they are and are not adjusted; Appendix~\ref{app:explore} looks at them in an exploratory analysis. Because
the denominator is all answers, the shares of correct answers of the models with mostly uninterpretable answers are not
comparable with those of \mname{Opus 5.5} and \mname{Sonnet 4.6}. Among interpretable answers (exploratory, after the
language-model review; Table~\ref{tab:interp}), \mname{Opus 5.5} was correct in 46.9\%, \mname{Sonnet 4.6} in 45.9\%,
\mname{NB-Llama 8B} in 27.2\% and \mname{Haiku 4.5} in 7.5\% of its 53 interpretable answers. Most wrong answers were
values not in the item's history, for example 359 of 408 for \mname{Opus 5.5} (Table~\ref{tab:wrong}). By value type
(FS3; counts and shares only), \mname{Opus 5.5} answered 70.8\% of the percentage items correctly against 32.9\% of
the NOK amounts, and \mname{Sonnet 4.6} 63.3\% against 27.5\% (Table~\ref{tab:types}).

\begin{table*}[t]
\centering
\footnotesize
\caption{Preregistered tests, primary analysis (Holm within each hypothesis \citep{holm1979}). n.t.\ = not tested: \textsuperscript{a}\,below the H1 minimum: $d$ and its interval are reported without a test, and with no outdated answer after the cutoff the interval is that of minus the share outdated before, conditional on drawing a series with an item after the cutoff (Section~\ref{sec:results}); it is not evidence against H1. \textsuperscript{b}\,below the H2 minimum; \textsuperscript{c}\,$b+c=0$. -- = not computed. Percentages and percentage points are rounded to one decimal, half to even.}
\label{tab:tests}
\begin{tabular}{llllll}
\toprule
\multicolumn{6}{l}{\textbf{H1}: $d$ = share outdated after minus before the cutoff (pp)} \\
Model & $n$ after / before & $d$ & CI & $p$ (one-sided) & $H_0$ rejected (Holm) \\
\midrule
\mname{Haiku 4.5} & 56 / 511 & 0.0 & [0.0, 0.0] & 1.0000 & no \\
\mname{Sonnet 4.6} & 9 / 558 & $-$9.1 & [$-$13.6, $-$5.1] & n.t.\textsuperscript{a} & -- \\
\mname{Opus 5.5} & 1 / 566 & $-$9.2 & [$-$13.5, $-$5.6] & n.t.\textsuperscript{a} & -- \\
\mname{NB-Llama 8B} & \multicolumn{5}{l}{not tested (no cutoff)} \\
\mname{NorMistral 7B} & \multicolumn{5}{l}{not tested (no cutoff)} \\
\midrule
\multicolumn{6}{l}{\textbf{H2}: $s$ = share of outdated answers equal to the most recent differing value (\%)} \\
Model & Outdated answers & $s$ & CI & $p$ (one-sided) & $H_0$ rejected (Holm) \\
\midrule
\mname{Haiku 4.5} & 0 & -- & -- & n.t.\textsuperscript{b} & -- \\
\mname{Sonnet 4.6} & 50 & 62.0 & [45.8, 81.4] & 0.0876 & no \\
\mname{Opus 5.5} & 46 & 50.0 & [31.4, 73.0] & 0.5027 & no \\
\mname{NB-Llama 8B} & 8 & 25.0 & [0.0, 42.9] & n.t.\textsuperscript{b} & -- \\
\mname{NorMistral 7B} & 0 & -- & -- & n.t.\textsuperscript{b} & -- \\
\bottomrule
\end{tabular}
\par\medskip
\begin{tabular}{lllllll}
\toprule
\multicolumn{7}{l}{\textbf{H3}: paired Bokmål/Nynorsk; $b$ = Bokmål correct and Nynorsk not, $c$ = the reverse} \\
Model & $b$ & $c$ & $p$ (exact McNemar) & Nynorsk $-$ Bokmål (pp) & CI & $H_0$ rejected (Holm) \\
\midrule
\mname{Haiku 4.5} & 1 & 0 & 0.5000 & $-$0.5 & [$-$1.6, 0.0] & no \\
\mname{Sonnet 4.6} & 9 & 3 & 0.0730 & $-$3.2 & [$-$8.6, 1.1] & no \\
\mname{Opus 5.5} & 3 & 8 & 0.9673 & 2.6 & [$-$1.0, 6.6] & no \\
\mname{NB-Llama 8B} & 6 & 1 & 0.0625 & $-$2.6 & [$-$6.7, 0.5] & no \\
\mname{NorMistral 7B} & 0 & 0 & n.t.\textsuperscript{c} & 0.0 & [0.0, 0.0] & -- \\
\bottomrule
\end{tabular}
\end{table*}

\paragraph{Tests.} Within each of H1, H2 and H3, the tests are corrected with the Holm--Bonferroni procedure
\citep{holm1979}; the sensitivity analyses report nominal $p$-values without correction. No power analysis was made in
advance, and none is made afterwards; instead, the text states what the intervals rule out. H4 of the preregistration,
that the share correct is lower for amounts and rates than for deadlines, was not tested in either run, because deadlines
were not included (\path{docs/PREREG_V2.md:32}).

\paragraph{H1.} H1 was tested only for \mname{Haiku 4.5}: 56 items in the H1 set took effect after its cutoff and 511
before, $d = 0.0$ percentage points and $p = 1.0000$ (Table~\ref{tab:tests}). \mname{Haiku 4.5} gave no outdated answer
in either group, so $d$ was 0 in every resample, and the test carries no information about H1. For
\mname{Sonnet 4.6} and \mname{Opus 5.5}, only 9 and 1 items took effect after the cutoff, below the minimum of 30 items
from at least 5 series, so $d$ is reported without a test. Neither model gave an outdated answer after its cutoff, so
each interval is that of minus the share outdated before, in the resamples that drew a series with an item after the
cutoff. For \mname{Opus 5.5}, 3\,696 of the 10\,000 resamples were dropped; its interval rests on the 6\,304 resamples
in which $d$ is defined. These intervals are not evidence against H1. The denominator of $d$ includes refusals: after
the cutoff, 8 of the 9 answers of \mname{Sonnet 4.6} and all 56 answers of \mname{Haiku 4.5} were uninterpretable,
against 74 of 558 and 475 of 511 before (Table~\ref{tab:h1groups}). Under the providers' reliable-knowledge cutoffs
(exploratory, after the language-model review), the groups after the cutoff would have 64, 36 and 1 items for \mname{Haiku 4.5}, \mname{Sonnet 4.6} and
\mname{Opus 5.5}; no test was run on them. \mname{NB-Llama 8B} and
\mname{NorMistral 7B} have no cutoff and were not tested. No model had a vague cutoff, so the rule in section 3.2 of the
run protocol (a conventional date and a sensitivity analysis without vague models) was not used; the addendum gives a
vague cutoff the value null and no H1 test.

\paragraph{H2.} \mname{Sonnet 4.6} gave 50 outdated answers in the H2 set, and $s = 62.0\%$ of them were the most
recent differing value ($p = 0.0876$). \mname{Opus 5.5} gave 46, with $s = 50.0\%$ ($p = 0.5027$). Neither null
hypothesis was rejected after Holm. The intervals rule out $s$ below 45.8\% for \mname{Sonnet 4.6} and below 31.4\%
for \mname{Opus 5.5}, and both include 50\%, so the test is inconclusive. \mname{Haiku 4.5}, \mname{NB-Llama 8B} and
\mname{NorMistral 7B} had fewer than 30 outdated answers and were not tested.

\paragraph{H3.} H3 uses an exact one-sided McNemar test \citep{mcnemar1947}. On the 189 pairs, no null
hypothesis was rejected after Holm. The lowest $p$-value was 0.0625 (\mname{NB-Llama
8B}, $b = 6$, $c = 1$). \mname{NorMistral 7B} had $b + c = 0$ and was not tested. An exact interval computed from $b$ and
$c$ alone (exploratory, after the language-model review; Table~\ref{tab:h3cond}) rules out a Nynorsk disadvantage larger than 5.7
percentage points for \mname{Sonnet 4.6} and 3.7 for \mname{NB-Llama 8B}; it ignores the clustering, and the
preregistered bootstrap intervals in Table~\ref{tab:tests} are wider. The test is inconclusive.

\begin{table}[t]
\centering
\footnotesize
\caption{FS2 (descriptive, no test): age in years of outdated answers, primary set.}
\label{tab:fs2}
\begin{tabular}{llllll}
\toprule
Model & Outdated & Q1 & Median & Q3 & CI for median \\
\midrule
\mname{Haiku 4.5} & 1 & 12.00 & 12.00 & 12.00 & [12.00, 12.00] \\
\mname{Sonnet 4.6} & 57 & 1.00 & 1.00 & 3.00 & [1.00, 2.00] \\
\mname{Opus 5.5} & 57 & 1.00 & 1.00 & 4.00 & [1.00, 3.00] \\
\mname{NB-Llama 8B} & 11 & 1.50 & 3.00 & 4.50 & [1.50, 5.00] \\
\mname{NorMistral 7B} & 0 & -- & -- & -- & -- \\
\bottomrule
\end{tabular}
\end{table}

\paragraph{FS2.} FS2 is descriptive, without a test (Table~\ref{tab:fs2}). The median age of outdated answers was 1
year for \mname{Sonnet 4.6} and \mname{Opus 5.5}, with 57 outdated answers each, and 3 years for \mname{NB-Llama 8B},
with 11. \mname{Haiku 4.5} gave one outdated answer, 12 years old.

\begin{table*}[t]
\centering
\footnotesize
\caption{Sensitivity analyses: nominal one-sided $p$-values, no Holm correction. S1 = all items and pairs; S2 = scorer v1 (H2 is not defined under v1); S3 = primary set without the flagged items. -- = no cutoff; n.t. = not tested (below a minimum or $b+c=0$).}
\label{tab:sensp}
\begin{tabular}{lllllllll}
\toprule
Model & S1 H1 & S1 H2 & S1 H3 & S2 H1 & S2 H3 & S3 H1 & S3 H2 & S3 H3 \\
\midrule
\mname{Haiku 4.5} & 1.0000 & n.t. & 0.5000 & 1.0000 & 0.5000 & 1.0000 & n.t. & 0.5000 \\
\mname{Sonnet 4.6} & 1.0000 & 0.0738 & 0.0730 & n.t. & 0.0730 & n.t. & 0.1018 & 0.0730 \\
\mname{Opus 5.5} & n.t. & 0.4157 & 0.9673 & n.t. & 0.9673 & n.t. & 0.5027 & 0.9673 \\
\mname{NB-Llama 8B} & -- & n.t. & 0.0625 & -- & 0.3438 & -- & n.t. & 0.0625 \\
\mname{NorMistral 7B} & -- & n.t. & n.t. & -- & n.t. & -- & n.t. & n.t. \\
\bottomrule
\end{tabular}
\end{table*}

\paragraph{Sensitivity analyses.} S1 uses all 916 items and 200 pairs, S2 scores the same answers with v1, and S3 is
the primary set without the 7 flagged items (873 items). They report nominal $p$-values without correction
(Table~\ref{tab:sensp}; class shares in Table~\ref{tab:sensclasses}). No nominal $p$-value is below 0.05. Scorer v1
counts later values as outdated and treats numbers that differ by less than 0.005 as equal; under v1 the outdated share
is higher, for \mname{Sonnet 4.6} 11.1\% against 6.5\%.

\section{Second run}\label{sec:v2}

\paragraph{Preregistration.} The second run was preregistered after the results of the first run and the
language-model review of the first version of this paper (Section~\ref{sec:ai}) were known. Its hypotheses and its item
set, the fourth lock, are those of the first run; the design was changed to make the hypotheses testable, and the models
were chosen by a criterion committed before the choice. The preregistration, the lock and the locked code were
published as Zenodo version 2.0 (DOI \texttt{10.5281/zenodo.23437444}) at 13:53:28 UTC on 17 October 2026, before any call
of the second run; its pilot started at 13:54:47 UTC. The analysis is the locked \texttt{analysis\_v2.py}, and the
results file RESULTS\_V2.md is rendered from it by a locked report script.

\paragraph{Models and runs.} Table~\ref{tab:v2models} lists the models of the locked list. In the pilot, every sample
of \mname{Gemini 2.5 Pro} and \mname{Gemini 2.5 Flash-Lite} ended in an error stating that the model is no longer
available to new users, and both were dropped (\mname{A-V2-3}); \mname{Gemini 2.5 Flash} was dropped because the free
quota could not cover its calls (\mname{A-V2-5}). The pilot's thinking rule gave the \mname{Gemma 4} models three runs each
(\mname{A-V2-9}); run 1 is primary. The eight remaining models were run on the main task (916 items) and on H3 (880
pairs, 1\,760 questions), with no system prompt, temperature 0 and at most 16\,384 output tokens for the API models and
1\,024 for the local models. Each model with a single run also had a control run on the main task for 100 fixed
primary items, used only to compare runs (\path{docs/PREREG_V2.md:186}). All logs have status success, every run was completed at the first attempt, and no run was
discarded or stopped. Final errors (samples still in error after the retries, scored uninterpretable; \path{docs/PREREG_V2.md:47},
\path{scripts/analysis_v2.py:93}) occurred only for
\mname{NorMistral 7B}: 3 in run 1 of the main task, 1 in its control run of 100 items and 4 on H3. In run 1 of the main
task, 349 answers of \mname{Gemma 4 26B}, 297 of \mname{NB-Llama 8B}, 10 of \mname{NorMistral 7B} and 8 of
\mname{Gemma 4 31B} reached the token limit and were scored as they stood. The API runs cost 30.35 USD including the pilot.

\begin{table*}[t]
\centering
\footnotesize
\caption{Second run: the model list as locked (PREREG\_V2.md, section 5) and the status of each model in the run. Deviations are listed in Table~\ref{tab:deviations}.}
\label{tab:v2models}
\begin{tabular}{llll}
\toprule
Inspect model id & Role & Cutoff & Status \\
\midrule
{\scriptsize\path{anthropic/claude-haiku-4-5-20251001}} & v1, H1 & 2025-07-31 & analysed \\
{\scriptsize\path{anthropic/claude-opus-4-5-20251101}} & H1 & 2025-08-31 & analysed \\
{\scriptsize\path{anthropic/claude-opus-4-6}} & H1 & 2025-08-31 & analysed \\
{\scriptsize\path{anthropic/claude-sonnet-4-6}} & v1 & 2026-01-31 & analysed \\
{\scriptsize\path{google/gemini-2.5-flash}} & H1 & 2025-01-31 & dropped (\mname{A-V2-5}) \\
{\scriptsize\path{google/gemini-2.5-flash-lite}} & H1 & 2025-01-31 & dropped (\mname{A-V2-3}) \\
{\scriptsize\path{google/gemini-2.5-pro}} & H1 & 2025-01-31 & dropped (\mname{A-V2-3}) \\
{\scriptsize\path{google/gemma-4-26b-a4b-it}} & H1 & 2025-01-31 & analysed (\mname{A-V2-9}) \\
{\scriptsize\path{google/gemma-4-31b-it}} & H1 & 2025-01-31 & analysed (\mname{A-V2-9}) \\
{\scriptsize\path{ollama/nb-llama-3.1-8b-instruct-cardtmpl:latest}} & v1 (local) & unknown & analysed \\
{\scriptsize\path{ollama/normistral-7b-warm-instruct-cardtmpl:latest}} & v1 (local) & unknown & analysed \\
\bottomrule
\end{tabular}
\end{table*}
\begin{table*}[t]
\centering
\footnotesize
\caption{Second run, primary set, run 1, scorer v2: share of all answers (\%) with cluster-bootstrap CI in brackets (RESULTS\_V2.md, T3).}
\label{tab:v2primary}
\begin{tabular}{llllll}
\toprule
Model & $n$ & Correct & Outdated & Wrong & Uninterpretable \\
\midrule
\mname{Haiku 4.5} & 880 & 0.5 [0.0, 1.3] & 0.1 [0.0, 0.4] & 6.4 [3.6, 9.3] & 93.1 [89.7, 96.1] \\
\mname{Opus 4.5} & 880 & 45.1 [35.2, 53.6] & 8.8 [5.8, 11.8] & 37.3 [29.7, 46.5] & 8.9 [6.8, 11.5] \\
\mname{Opus 4.6} & 880 & 43.9 [34.1, 52.3] & 11.2 [8.0, 14.5] & 44.4 [36.0, 54.7] & 0.5 [0.1, 1.0] \\
\mname{Sonnet 4.6} & 880 & 39.5 [28.9, 48.9] & 6.7 [3.5, 10.1] & 40.9 [32.7, 50.6] & 12.8 [9.3, 17.5] \\
\mname{Gemma 4 26B} & 880 & 7.6 [3.8, 11.8] & 4.0 [1.7, 6.7] & 50.0 [44.5, 55.3] & 38.4 [32.9, 44.3] \\
\mname{Gemma 4 31B} & 880 & 11.7 [6.0, 18.4] & 5.5 [2.9, 8.6] & 81.6 [74.1, 88.7] & 1.2 [0.5, 2.1] \\
\mname{NB-Llama 8B} & 880 & 8.5 [3.4, 14.3] & 1.2 [0.3, 2.4] & 21.6 [15.0, 28.3] & 68.6 [59.8, 78.1] \\
\mname{NorMistral 7B} & 880 & 0.0 [0.0, 0.0] & 0.0 [0.0, 0.0] & 0.2 [0.0, 0.6] & 99.8 [99.4, 100.0] \\
\bottomrule
\end{tabular}
\end{table*}

\paragraph{Primary measures.} On the primary set (880 items, 96 series; Table~\ref{tab:v2primary}), \mname{Opus 4.5},
\mname{Opus 4.6} and \mname{Sonnet 4.6} answered 45.1\%, 43.9\% and 39.5\% of the items correctly, \mname{Gemma 4 31B}
11.7\%, \mname{NB-Llama 8B} 8.5\%, \mname{Gemma 4 26B} 7.6\%, \mname{Haiku 4.5} 0.5\% and \mname{NorMistral 7B} 0.0\%.
Of the 880 answers of each model on the primary set, 93.1\% were uninterpretable for \mname{Haiku 4.5}, 99.8\% for
\mname{NorMistral 7B}, 68.6\% for \mname{NB-Llama 8B} and 38.4\% for \mname{Gemma 4 26B}; \mname{Gemma 4 31B} gave 81.6\% wrong answers.

\paragraph{H1: decision rule.} The preregistration fixes the primary test of H1 as a test pooled over the H1 models
(\path{docs/PREREG_V2.md:234} to \path{docs/PREREG_V2.md:237}; translated): $d$ is the share outdated after the cutoff
minus the share outdated before, each among interpretable answers and with the counts summed over the models; the null
hypothesis is $d \le 0$, tested one-sided at $\alpha = 0.05$; the test requires 30 interpretable answers after the
cutoff from at least 20 series, and otherwise $d$ is reported with an interval and without a test. Per model, the test
requires 30 interpretable answers in each group and 5 series after the cutoff, with Holm over the tested models
(\path{docs/PREREG_V2.md:239}). A test is \emph{supported} when the null hypothesis is rejected; \emph{inconclusive} when
it is not rejected and the 95\% interval reaches the smallest effect of interest, 10 percentage points for H1; and
\emph{no effect of interest} when the interval excludes it (\path{docs/PREREG_V2.md:283} to
\path{docs/PREREG_V2.md:291}). Sensitivity analysis S8 is defined in \path{docs/PREREG_V2.md:73}: «S8 (H1′) teller bare
svar som er lik den nyeste avvikende verdien. Da er målmengden én verdi i begge gruppene.» (S8 counts only answers equal
to the most recent value that differs from the key; the target set is then one value in both groups.) It was
preregistered because items after the cutoff have more earlier differing values than items before, so that a guessed
number is slightly more likely to be classed as outdated after the cutoff, and it uses nominal $p$-values
(\path{docs/PREREG_V2.md:266}).

\begin{table*}[t]
\centering
\scriptsize
\caption{Second run, H1 (RESULTS\_V2.md, T8). $d$ is the share outdated after minus before the cutoff among interpretable answers, in percentage points, with a stratified cluster-bootstrap 95\% interval; $p$ is one-sided. The pooled primary test decides H1; per-model decisions use Holm. S8 is a preregistered sensitivity analysis with nominal $p$. A model is tested only with at least 30 interpretable answers in each group.}
\label{tab:v2h1}
\begin{tabular}{llllllll}
\toprule
Model & $n$ after & Series after & $n$ before & $d$ & 95\% interval & $p$ & State \\
\midrule
\multicolumn{8}{l}{\emph{Primary analysis (H1), interpretable answers}} \\
Pooled & 115 & 55 & 1877 & 17.8 & [7.7, 28.9] & 0.0001 & supported \\
\mname{Haiku 4.5} & 0 & 0 & 40 & -- & -- & -- & not tested (Holm: --) \\
\mname{Opus 4.5} & 6 & 6 & 492 & 36.4 & [$-$12.4, 85.3] & -- & not tested (Holm: --) \\
\mname{Opus 4.6} & 35 & 35 & 528 & 34.2 & [17.5, 50.9] & 0.0001 & supported (Holm: yes) \\
\mname{Gemma 4 26B} & 15 & 13 & 318 & 1.0 & [$-$8.4, 17.9] & -- & not tested (Holm: --) \\
\mname{Gemma 4 31B} & 59 & 53 & 499 & 13.2 & [4.4, 23.2] & 0.0007 & supported (Holm: yes) \\
\midrule
\multicolumn{8}{l}{\emph{S8 (H1$'$): only the most recent value that differs from the key counts as outdated}} \\
Pooled & 115 & 55 & 1877 & 3.9 & [$-$3.6, 12.7] & 0.1754 & inconclusive \\
\mname{Haiku 4.5} & 0 & 0 & 40 & -- & -- & -- & not tested (nominal $p$: --) \\
\mname{Opus 4.5} & 6 & 6 & 492 & 7.1 & [$-$11.9, 41.9] & -- & not tested (nominal $p$: --) \\
\mname{Opus 4.6} & 35 & 35 & 528 & 7.5 & [$-$4.9, 21.8] & 0.1347 & inconclusive (nominal $p$: no) \\
\mname{Gemma 4 26B} & 15 & 13 & 318 & 3.8 & [$-$4.2, 20.4] & -- & not tested (nominal $p$: --) \\
\mname{Gemma 4 31B} & 59 & 53 & 499 & 4.0 & [$-$2.2, 11.6] & 0.1236 & inconclusive (nominal $p$: no) \\
\bottomrule
\end{tabular}
\end{table*}

\paragraph{H1: results.} Pooled over the five H1 models, 115 interpretable answers from 55 series fell after the cutoff
and 1\,877 before. The primary test rejected the null hypothesis, $d = 17.8$ percentage points, 95\% interval
[7.7, 28.9], $p = 0.0001$ (Table~\ref{tab:v2h1}), so H1 is supported under the decision rule. In S8, which uses the same
answers but one target value in both groups, $d = 3.9$ [$-$3.6, 12.7], $p = 0.1754$, inconclusive: the null hypothesis is
not rejected and the interval reaches the smallest effect of interest of 10 percentage points
(\path{docs/PREREG_V2.md:289}); it also contains 0. H1 was supported in all other preregistered sensitivity analyses (Table~\ref{tab:v2sens}), but none of
them gives both groups one target value; under S8, which does, the effect falls from 17.8 to 3.9 percentage points and is
not established. We therefore do not regard the support for H1 as robust. Per model,
\mname{Opus 4.6} ($d = 34.2$ [17.5, 50.9], $p = 0.0001$) and \mname{Gemma 4 31B} ($d = 13.2$ [4.4, 23.2],
$p = 0.0007$) were supported after Holm; in S8 they gave $d = 7.5$ [$-$4.9, 21.8], $p = 0.1347$, and $d = 4.0$
[$-$2.2, 11.6], $p = 0.1236$, both inconclusive. \mname{Haiku 4.5}, \mname{Opus 4.5} and \mname{Gemma 4 26B} had 0, 6
and 15 interpretable answers after their cutoffs and were not tested. On average, items after the cutoff had 7.72 to
9.61 earlier differing values and items before 6.20 to 6.27, depending on the model (RESULTS\_V2.md, T8).

\begin{table*}[t]
\centering
\scriptsize\setlength{\tabcolsep}{4pt}
\caption{Second run: all preregistered sensitivity analyses of H1, pooled over the H1 models, with nominal one-sided $p$-values (RESULTS\_V2.md, T16; S8 from T8). The definitions are quoted verbatim from the preregistration (PREREG\_V2.md, section 7.5). $d$ in percentage points with a 95\% interval. $p$ is one-sided and the interval two-sided, so a test can reject at 0.05 while its interval contains 0 (S9).}
\label{tab:v2sens}
\begin{tabular}{lp{7.2cm}llll}
\toprule
Analysis & Definition & $d$ & 95\% interval & $p$ & State \\
\midrule
S1 & «Alle 916 items; H3 har de samme 880 parene» ({\scriptsize\path{docs/PREREG_V2.md:271}}) & 10.6 & [2.9, 19.7] & 0.0032 & supported \\
S2 & «Scorer v1 (H1 og H3; H2 er ikke definert under v1)» ({\scriptsize\path{docs/PREREG_V2.md:272}}) & 11.9 & [2.2, 22.6] & 0.0071 & supported \\
S3 & «Uten de 7 flaggede itemene» ({\scriptsize\path{docs/PREREG_V2.md:273}}) & 18.4 & [8.2, 29.3] & 0.0005 & supported \\
S4 & «H1 med alle svar som nevner» ({\scriptsize\path{docs/PREREG_V2.md:274}}) & 5.2 & [0.7, 10.0] & 0.0115 & supported \\
S5 & «H1 med «reliable knowledge cutoff» der leverandøren oppgir den, ellers samme cutoff» ({\scriptsize\path{docs/PREREG_V2.md:275}}) & 14.0 & [5.0, 24.0] & 0.0007 & supported \\
S6, run 2 & «Samme analyse på kjøring 2 og 3 for modellene med R = 3 som er komplette i den kjøringen» ({\scriptsize\path{docs/PREREG_V2.md:276}}) & 10.7 & [3.5, 19.5] & 0.0009 & supported \\
S6, run 3 & as above & 10.7 & [3.5, 19.5] & 0.0009 & supported \\
S7 & «Uten itemene i `data/errata\_v2.txt` (lik primæranalysen hvis fila er tom)» ({\scriptsize\path{docs/PREREG_V2.md:277}}) & 17.8 & [7.7, 28.9] & 0.0001 & supported \\
S8 (T8) & «H1′: bare svar lik den nyeste avvikende verdien teller som utdatert» ({\scriptsize\path{docs/PREREG_V2.md:278}}) & 3.9 & [$-$3.6, 12.7] & 0.1754 & inconclusive \\
S9 & «H1 med gruppen «før» begrenset til items som tok virkning de siste 24 månedene før cutoff. Den skiller cutoff fra bestemmelsens alder: items etter cutoff er de nyeste.» ({\scriptsize\path{docs/PREREG_V2.md:279}}) & 8.5 & [$-$0.7, 18.4] & 0.0359 & supported \\
\bottomrule
\end{tabular}
\end{table*}

\begin{table*}[t]
\centering
\footnotesize
\caption{Second run, H1 set: answers after and before each H1 model's cutoff. Items and uninterpretable answers are from RESULTS\_V2.md, T19; interpretable answers are the $n$ of the primary H1 analysis (T8). Uninterpretable is the scorer-v2 class \emph{utolket}: no answer line, no number, or more than one number.}
\label{tab:v2interp}
\begin{tabular}{lllllll}
\toprule
& \multicolumn{3}{l}{After the cutoff} & \multicolumn{3}{l}{Before the cutoff} \\ Model & Items & Interpretable & Uninterpretable & Items & Interpretable & Uninterpretable \\
\midrule
\mname{Haiku 4.5} & 56 & 0 & 56 & 511 & 40 & 471 \\
\mname{Opus 4.5} & 36 & 6 & 30 & 531 & 492 & 39 \\
\mname{Opus 4.6} & 36 & 35 & 1 & 531 & 528 & 3 \\
\mname{Gemma 4 26B} & 64 & 15 & 49 & 503 & 318 & 185 \\
\mname{Gemma 4 31B} & 64 & 59 & 5 & 503 & 499 & 4 \\
\bottomrule
\end{tabular}
\end{table*}

\paragraph{Interpretable answers around the cutoff.} Table~\ref{tab:v2interp} gives the answers after and before each
H1 model's cutoff. Uninterpretable is the scorer-v2 class \emph{utolket}, which the scorer defines as ``no answer line,
no number, or more than one number'' (\path{src/norregel/score_v2.py:22}). \mname{Haiku 4.5} gave an uninterpretable
answer to all 56 items after its cutoff and to 471 of the 511 before. \mname{Opus 4.5} gave 30 of 36 after and 39 of
531 before. \mname{Opus 4.6} gave 1 of 36 after and 3 of 531 before. \mname{Gemma 4 26B} gave 49 of 64 after and 185
of 503 before. \mname{Gemma 4 31B} gave 5 of 64 after and 4 of 503 before.

\begin{table}[t]
\centering
\footnotesize
\caption{Second run, H2 as estimation (RESULTS\_V2.md, T9): $s$ is the share (\%) of outdated answers that equal the most recent value that differs from the key, with a cluster-bootstrap 95\% interval; no test.}
\label{tab:v2h2}
\begin{tabular}{llllll}
\toprule
Model & Outdated & Series & $s$ & 95\% interval & Interval vs 0.5 \\
\midrule
Pooled & 287 & 33 & 56.1 & [46.6, 65.0] & contains 0.5 \\
\mname{Haiku 4.5} & 0 & 0 & fewer than 30 & -- & -- \\
\mname{Opus 4.5} & 70 & 23 & 65.7 & [50.5, 81.0] & above 0.5 \\
\mname{Opus 4.6} & 88 & 30 & 58.0 & [46.6, 68.9] & contains 0.5 \\
\mname{Sonnet 4.6} & 52 & 19 & 53.8 & [38.2, 75.8] & contains 0.5 \\
\mname{Gemma 4 26B} & 30 & 12 & 50.0 & [20.7, 78.6] & contains 0.5 \\
\mname{Gemma 4 31B} & 39 & 15 & 48.7 & [24.0, 65.0] & contains 0.5 \\
\mname{NB-Llama 8B} & 8 & 4 & fewer than 30 & -- & -- \\
\mname{NorMistral 7B} & 0 & 0 & fewer than 30 & -- & -- \\
\bottomrule
\end{tabular}
\end{table}

\paragraph{H2.} H2 is reported as an estimate of $s$, the share of outdated answers that equal the most recent value
that differs from the key, with no test (\path{docs/PREREG_V2.md:251}). Pooled over all models, 287 outdated answers
from 33 series gave $s = 56.1\%$ [46.6, 65.0], an interval that contains 0.5 (Table~\ref{tab:v2h2}). Of the models with
at least 30 outdated answers, the interval lies above 0.5 for \mname{Opus 4.5} ($s = 65.7\%$ [50.5, 81.0]) and contains
0.5 for \mname{Opus 4.6}, \mname{Sonnet 4.6}, \mname{Gemma 4 26B} and \mname{Gemma 4 31B}.

\begin{table*}[t]
\centering
\scriptsize\setlength{\tabcolsep}{3pt}
\caption{Second run, H3 (RESULTS\_V2.md, T10): $b$ pairs with only Bokmål correct, $c$ pairs with only Nynorsk correct; $D = (c - b)/\mathrm{pairs}$ in percentage points with a cluster-bootstrap 95\% interval; $p$ is one-sided for a lower share correct in Nynorsk. Per-model decisions use the exact McNemar test with Holm. The last column is the share of uninterpretable answers among all answers of the H3 task, run 1 (descriptive count written after the analysis, anc/H3\_UTOLKET\_V2.md).}
\label{tab:v2h3}
\begin{tabular}{lllllllll}
\toprule
Model & $b$ & $c$ & Pairs & $D$ & 95\% interval & $p$ & State & Uninterpretable (\%) \\
\midrule
Pooled & 173 & 146 & 7040 & $-$0.4 & [$-$1.1, 0.3] & 0.1495 & inconclusive & -- \\
\mname{Haiku 4.5} & 11 & 0 & 880 & $-$1.2 & [$-$3.6, 0.0] & 0.0005 & supported (Holm: yes) & 93.4 \\
\mname{Opus 4.5} & 24 & 35 & 880 & 1.2 & [$-$0.9, 3.4] & 0.9413 & no effect of interest (Holm: no) & 8.4 \\
\mname{Opus 4.6} & 14 & 26 & 880 & 1.4 & [$-$0.2, 3.2] & 0.9808 & no effect of interest (Holm: no) & 0.3 \\
\mname{Sonnet 4.6} & 39 & 26 & 880 & $-$1.5 & [$-$4.0, 0.7] & 0.0680 & inconclusive (Holm: no) & 12.6 \\
\mname{Gemma 4 26B} & 16 & 21 & 880 & 0.6 & [$-$1.1, 2.4] & 0.8380 & inconclusive (Holm: no) & 37.3 \\
\mname{Gemma 4 31B} & 8 & 20 & 880 & 1.4 & [0.1, 2.9] & 0.9937 & no effect of interest (Holm: no) & 1.4 \\
\mname{NB-Llama 8B} & 59 & 18 & 880 & $-$4.7 & [$-$10.9, 0.3] & $<$0.0001 & supported (Holm: yes) & 68.3 \\
\mname{NorMistral 7B} & 2 & 0 & 880 & $-$0.2 & [$-$0.6, 0.0] & 0.2500 & no effect of interest (Holm: no) & 99.7 \\
\bottomrule
\end{tabular}
\end{table*}

\paragraph{H3.} The pooled test of a lower share correct in Nynorsk than in Bokmål gave $b = 173$ and $c = 146$ over
7\,040 pairs, $D = -0.4$ percentage points [$-$1.1, 0.3], $p = 0.1495$: inconclusive, because the interval reaches the
smallest effect of interest of 1 percentage point (Table~\ref{tab:v2h3}). Per model, with the exact McNemar test and Holm, H3 was
supported for \mname{Haiku 4.5} ($b = 11$, $c = 0$, $p = 0.0005$) and for \mname{NB-Llama 8B} ($b = 59$, $c = 18$,
$p < 0.0001$) under the decision rule. \mname{Sonnet 4.6} ($b = 39$, $c = 26$, $p = 0.0680$) and \mname{Gemma 4 26B}
($b = 16$, $c = 21$) were inconclusive, and \mname{Opus 4.5} ($b = 24$, $c = 35$), \mname{Opus 4.6} ($b = 14$,
$c = 26$), \mname{Gemma 4 31B} ($b = 8$, $c = 20$) and \mname{NorMistral 7B} ($b = 2$, $c = 0$) showed no effect of
interest. Of the 1\,760 answers of each model in the H3 task, 93.4\% were uninterpretable for \mname{Haiku 4.5}, 68.3\%
for \mname{NB-Llama 8B}, 99.7\% for \mname{NorMistral 7B}, 37.3\% for \mname{Gemma 4 26B}, 12.6\% for \mname{Sonnet 4.6},
8.4\% for \mname{Opus 4.5}, 1.4\% for \mname{Gemma 4 31B} and 0.3\% for \mname{Opus 4.6} (Table~\ref{tab:v2h3}, last
column; a descriptive count made after the analysis, anc/H3\_UTOLKET\_V2.md).

\paragraph{A caution rule set after the analysis.} After the analysis, and not as part of the preregistration, the
author introduced a caution rule: a model for which more than half of the answers in the H3 task could not be
interpreted is reported with its numbers and is not interpreted, whatever the outcome. By Table~\ref{tab:v2h3} the rule
applies to \mname{Haiku 4.5}, \mname{NB-Llama 8B} and \mname{NorMistral 7B}, so the results for which H3 was supported
for \mname{Haiku 4.5} and \mname{NB-Llama 8B}, and the result of \mname{NorMistral 7B}, are not interpreted. The result
of \mname{Haiku 4.5} also rests on 11 discordant pairs.

\section{Deviations and amendments}\label{sec:dev}

Table~\ref{tab:deviations} lists every deviation and amendment of both runs with its source. The first run's
deviations D1--D12 and amendments A1--A8 are in the run protocol, with date and reason, and deviation D13 is recorded in
the results. The second run's \mname{A-V2-1} and \mname{A-V2-2} were made before locking and are recorded in the
preregistration; \mname{A-V2-3} to \mname{A-V2-9} were made after locking and are recorded in AVVIK\_V2.md.
\mname{A-V2-9} is a triggered rule: the locked pilot rule gives a \mname{Gemma 4} model three runs when its pilot log counts
thinking tokens (\path{docs/PREREG_V2.md:177}), and the pilot counted 170\,567 and 26\,924 for \mname{Gemma 4 26B} and
\mname{Gemma 4 31B}; the preregistration requires triggered rules to be recorded as deviations
(\path{docs/PREREG_V2.md:181}). The H3 runs of the second run used a run script whose path to the H3 items was
corrected after locking (\mname{A-V2-6}), and they ran under the corrected lock before it was published
(\mname{A-V2-7}). Errors in the published text of version 2.0 are listed in ERRATA\_V2.md.

\begin{table*}[t]
\centering
\footnotesize
\caption{All deviations and amendments of the first (D, A) and second run (A-V2). The source is the repository file and line; D13 is recorded in the results file.}
\label{tab:deviations}
\begin{tabular}{lp{9.5cm}l}
\toprule
Code & Content & Source \\
\midrule
\mname{D1} & the log directory differs from the addendum's command & {\scriptsize\path{docs/RUN_PROTOCOL.md:211}} \\
\mname{D2} & the local models ran with a single connection & {\scriptsize\path{docs/RUN_PROTOCOL.md:212}} \\
\mname{D3} & the years-and-months rule is applied in the analysis, not in the locked scorer & {\scriptsize\path{docs/RUN_PROTOCOL.md:214}} \\
\mname{D4} & the provider client libraries were installed in the existing environment & {\scriptsize\path{docs/RUN_PROTOCOL.md:215}} \\
\mname{A1} & replaced client library versions, before any request & {\scriptsize\path{docs/RUN_PROTOCOL.md:216}} \\
\mname{D5} & the model list was bound by a local commit during the runs & {\scriptsize\path{docs/RUN_PROTOCOL.md:217}} \\
\mname{D6} & \mname{NorMistral} ran with the chat template of its model card & {\scriptsize\path{docs/RUN_PROTOCOL.md:218}} \\
\mname{A2} & amendment behind D6 & {\scriptsize\path{docs/RUN_PROTOCOL.md:219}} \\
\mname{D7} & \mname{NB-Llama} ran with the card's template with a fixed date, on its own Ollama instance & {\scriptsize\path{docs/RUN_PROTOCOL.md:226}} \\
\mname{A3} & amendment behind D7 & {\scriptsize\path{docs/RUN_PROTOCOL.md:227}} \\
\mname{D8} & \mname{NorMistral} was to be kept out of the comparisons & {\scriptsize\path{docs/RUN_PROTOCOL.md:240}} \\
\mname{A4} & amendment behind D8, repealed by A5 & {\scriptsize\path{docs/RUN_PROTOCOL.md:241}} \\
\mname{D9} & \mname{NorMistral} was run complete and is reported like the other models & {\scriptsize\path{docs/RUN_PROTOCOL.md:247}} \\
\mname{A5} & amendment behind D9 & {\scriptsize\path{docs/RUN_PROTOCOL.md:248}} \\
\mname{D10} & \mname{NorMistral} ran with a finite number of provider-level retries & {\scriptsize\path{docs/RUN_PROTOCOL.md:254}} \\
\mname{A6} & amendment behind D10 & {\scriptsize\path{docs/RUN_PROTOCOL.md:255}} \\
\mname{A7} & editorial corrections before the publication of v1.1 & {\scriptsize\path{docs/RUN_PROTOCOL.md:262}} \\
\mname{D11} & local and API runs overlapped & {\scriptsize\path{docs/RUN_PROTOCOL.md:295}} \\
\mname{D12} & the model list was published after the runs and the analysis & {\scriptsize\path{docs/RUN_PROTOCOL.md:296}} \\
\mname{A8} & corrections after proofreading before v1.1 & {\scriptsize\path{docs/RUN_PROTOCOL.md:297}} \\
\mname{D13} & API runs overlapped & {\scriptsize\path{docs/RESULTS.md:9}} \\
\mname{A-V2-1} & \mname{Opus 5.5} was removed before locking & {\scriptsize\path{docs/PREREG_V2.md:445}} \\
\mname{A-V2-2} & the rule on the Google data terms was made stricter, before locking & {\scriptsize\path{docs/PREREG_V2.md:455}} \\
\mname{A-V2-3} & \mname{Gemini 2.5 Pro} and \mname{Gemini 2.5 Flash-Lite} dropped: not available to new users in the pilot & {\scriptsize\path{docs/AVVIK_V2.md:13}} \\
\mname{A-V2-4} & a run in which every sample is an error counts as dropped & {\scriptsize\path{docs/AVVIK_V2.md:26}} \\
\mname{A-V2-5} & \mname{Gemini 2.5 Flash} dropped: the free quota could not cover the calls & {\scriptsize\path{docs/AVVIK_V2.md:33}} \\
\mname{A-V2-6} & the path of the H3 task in the run script corrected; switch ONLY\_H3 added & {\scriptsize\path{docs/AVVIK_V2.md:45}} \\
\mname{A-V2-7} & the H3 runs were made under the corrected lock before it was published & {\scriptsize\path{docs/AVVIK_V2.md:65}} \\
\mname{A-V2-8} & the pilot logs are not listed in table T2 & {\scriptsize\path{docs/AVVIK_V2.md:72}} \\
\mname{A-V2-9} & the thinking rule of the pilot was triggered: \mname{Gemma 4} got more runs (triggered rule) & {\scriptsize\path{docs/AVVIK_V2.md:78}} \\
\bottomrule
\end{tabular}
\end{table*}

The first run's deviations in more detail:
\begin{itemize}\itemsep0pt
\item D1: the log directory differs from the addendum's command.
\item D2: the local models ran with one connection, not 10.
\item D3: the «år og måneder» (years and months) rule is applied in the analysis, not in the locked scorer.
\item D4 and A1: the provider client libraries were installed in the existing environment
  (\texttt{anthropic 0.125.0}, \texttt{openai 2.54.0}); A1 replaced versions that Inspect could not use, before any
  request was sent.
\item D5: during the runs the model list was bound by a local commit, not by a published version (see D12).
\item D6 and A2: \mname{NorMistral 7B} ran with the chat template of its model card.
\item D7 and A3: \mname{NB-Llama 8B} ran with the card's template with a fixed date, on its own Ollama instance. A3
  followed a reading of the answers of the earlier \mname{NB-Llama 8B} run, in which 48 of the 99 samples stored when the answers were read had reached the limit of 1\,024 tokens (109 had been stored when the run was stopped); that run was discarded before analysis.
\item D8 and A4: \mname{NorMistral 7B} was to be kept out of all comparisons after all 20 pilot answers were
  uninterpretable. A5 repealed A4.
\item D9 and A5: \mname{NorMistral 7B} was run complete and is reported like the other models.
\item D10 and A6: \mname{NorMistral 7B} ran with 2 provider-level retries after Ollama answered a request with HTTP 500
  on every attempt and Inspect retried without end.
\item D11: the addendum says «one model at a time». The first \mname{NB-Llama 8B} run (16:45:40--17:42:24 UTC,
  discarded before analysis) ran in parallel with the API models. The analysed local runs started after the last API
  run had finished (17:19:17 UTC), so no analysed local run overlapped with an API run. This makes the description of
  D11 in the run protocol (v1.1) more precise.
\item \textbf{D12}: the model list was published after the runs and the analysis had been completed locally, and
  before any result was made public; the addendum requires publication before any result (Section~\ref{sec:prereg}).
\item D13: the run protocol (v1.1) says that the API models were run one at a time, but the \mname{Sonnet 4.6} H3 run
  (16:55:33--16:58:02 UTC) and the \mname{Opus 5.5} main run (start 16:56:09 UTC) overlapped on 16 October 2026 (times
  from the log headers).
\item A7 and A8: editorial corrections, made after the results were known locally but before they were made
  public, and before the publication of v1.1, with no change in method, models, parameters or analysis.
\end{itemize}

\section{Limitations}

\paragraph{Recorded before the runs.} In 73 items in 15 series, the anchor phrase in the question is not verbatim.
In 27 groups of items (69 items), the same value, source and date appear in different paragraphs, and the bootstrap treats
them as different series. In H3, document titles and anchor phrases keep the written standard of the source, so a
Nynorsk question can contain Bokmål words and the other way round; the contrast concerns only the question frame.
Language-model sub-agents directed by the author took part in the cross-checks and in the audit of the fourth lock. The
addendum also left open the cutoff rule, vague cutoffs, the binding of the model list, restarts, a conflict between
scorer rules, text against code, the two meanings of «primary set», details of sensitivity analysis S2, token and
temperature limits, library versions and part of the audit. The run protocol settled several of these before the
runs.

\paragraph{Revealed by the first run.} (i) \mname{Haiku 4.5}, \mname{NB-Llama 8B} and \mname{NorMistral 7B} have large
uninterpretable shares. For \mname{NorMistral 7B}, a possible contributing factor is that its chat markers are stored
as ordinary tokens in the model file; this has not been shown (A5). (ii) H1 could not be tested in substance:
the only test had no outdated answer in either group, \mname{Sonnet 4.6} and \mname{Opus 5.5} had too few items after their cutoffs,
and the local models have no cutoff. (iii) The cutoffs are those the providers state; effective cutoffs can differ
\citep{cheng2024dated}. In its answers, \mname{Haiku 4.5} gave April 2024 as the end of its knowledge in 69 answers,
and \mname{Sonnet 4.6} gave early 2025 in 2 answers and July 2025 in 1. (iv) Only the discarded first \mname{NB-Llama 8B} run ran in parallel with the API models, and no analysed local run
overlapped with an API run (D11); runs of different API models overlapped (D13), and the model list was
published after the analysis had been completed locally (D12). (v) In 234 items of the primary set some earlier value equals the key (in 219 of them the preceding value), so in
these items an outdated answer cannot be told from a correct answer (\path{docs/PREREG_TILLEGG.md:70}). (vi) All three API models come from the same provider, Anthropic, whose tool Claude Code was also used in this work (Section~\ref{sec:ai}). (vii) The only H1 test was run on a model that answered 0.5\% of the primary items correctly and gave 94.0\% uninterpretable answers among its 880 answers on the primary set.

\paragraph{Constructs and measurement.} (viii) An answer is outdated only when it equals an earlier value in the clean
series. A value from before a series split counts as not in the history, so that class is an upper bound for values
that never applied. In the first run, no estimate was made of how often a guessed value equals an earlier value by
chance, which matters most for percentages; the second run estimated it (table T7 in RESULTS\_V2.md,
\path{docs/RESULTS_V2.md:111}). In both runs an outdated answer cannot be told from a known value given for the wrong date. (ix) In the first run,
all shares and $d$ use all answers as the denominator (for $d$, all answers in the group;
\path{docs/PREREG_TILLEGG.md:258}), so refusals and unreadable answers count as neither correct nor outdated;
Table~\ref{tab:interp} gives the shares among interpretable answers. In the second run, the class shares also use all
answers, but $d$ is computed among interpretable answers (\path{docs/PREREG_V2.md:235}). (x) In the first run, \mname{Opus 5.5} ran with
thinking, with up to 16\,384 tokens and without a temperature setting, the other models with at most 1\,024 tokens and
temperature 0, and each model ran once, so the configurations differ and variation between runs is not measured. The
second run measured it for all eight models (table T17 in RESULTS\_V2.md, \path{docs/RESULTS_V2.md:355}): the
\mname{Gemma 4} models gave the same class as in run 1 to all 880 primary items in runs 2 and 3, and in the control
runs on 100 primary items the other models gave the same class as in run 1 to between 94.0\% (\mname{Sonnet 4.6}) and
100.0\% (\mname{NB-Llama 8B} and \mname{NorMistral 7B}) of the items. (xi) In the
final check of the fourth lock, 37 of the 40 checked items are in the primary set, with no wrong key and no wrong
value; the 95\% Clopper--Pearson upper bound for a wrong key is 9.5\%, which corresponds to about 80 of the 843
primary items that were not checked. S3 removes only the 7 known flagged items and does not cover unknown errors.

\paragraph{Second run.} (xii) The H1 effect of the primary test is not established in S8 (Section~\ref{sec:v2}), the only sensitivity analysis that
gives both groups one target value; all other preregistered sensitivity analyses support it (Table~\ref{tab:v2sens}). Items
after the cutoff have more earlier differing values than items before, and the primary measure does not separate this
from an effect of the cutoff. (xiii) Of the H1 models, only \mname{Opus 4.6} and \mname{Gemma 4 31B} met the per-model
minimum, and \mname{Haiku 4.5} gave no interpretable answer after its cutoff. (xiv) The \mname{Gemini 2.5} models were dropped
(\mname{A-V2-3}, \mname{A-V2-5}), so the H1 models are \mname{Haiku 4.5}, \mname{Opus 4.5}, \mname{Opus 4.6} and the
\mname{Gemma 4} models. (xv) The cutoffs are those the providers state; for the \mname{Gemma 4} models it is the date of the pretraining
data in the model card. (xvi) The second run used the same items as the first, whose results were known when it was
preregistered. (xvii) The uninterpretable shares of the H3 task were counted after the analysis.

\section{Discussion and conclusion}\label{sec:discussion}

\paragraph{What the results show.} In the first run, \mname{Opus 5.5} and \mname{Sonnet 4.6} gave an interpretable
answer to 876 and 771 of the 880 primary items. About half of their interpretable answers were correct. Outdated
answers made up 6.5\% of all answers for both models, and most wrong answers were values not in the item's history.
Percentages were answered correctly more often than NOK amounts. In the second run, \mname{Opus 4.5}, \mname{Opus 4.6}
and \mname{Sonnet 4.6} answered 45.1\%, 43.9\% and 39.5\% of the primary items correctly. In both runs, several models
gave mostly uninterpretable answers in this setup. The pooled primary test of H1 in the second run rejected its null
hypothesis, so H1 is supported under the preregistered decision rule. In S8, the preregistered control in which both
groups have one target value, the interval [$-$3.6, 12.7] contains 0. H1 was supported in every other preregistered sensitivity
analysis, but S8 is the only one that gives both groups one target value. We therefore do not regard the support for H1
as robust.

\paragraph{What they do not show.} The results do not establish that models give outdated values more often for rules that
changed after their cutoff. The first run could not test H1 in substance, and in the second run the effect of the
primary test, which every other preregistered sensitivity analysis supported, was not established once the target sets
were made equal. The S8 interval reaches both 0 and the smallest
effect of interest of 10 percentage points, so it neither rules the effect out nor supports it. The pooled estimate of
H2 contains 0.5, so a preference for the most recent value that differs from the key is neither shown nor ruled out.
The pooled H3 test was inconclusive. H3 was supported per model for \mname{Haiku 4.5} and \mname{NB-Llama 8B}, but
these results fall under the caution rule set after the analysis (more than half of the answers in the H3 task
uninterpretable) and are not interpreted. The first run did not estimate chance matches, and values from before a
series split count as not in the history, so the share of values that never applied is an upper bound, not a
measurement.

\section{Data and code availability}

The items (questions and reference values) are withheld to limit contamination of future training data; the item
files carry a canary string. The preregistration does not regulate access to the items. They are not distributed;
evaluation of further models against the held-out items can be arranged with EcoDeco AS. Item IDs, the
exclusion list and per-item score classes without reference values are published with the results. The run logs contain
the keys in their metadata and are not published. The Zenodo record has the versions v1 (\path{10.5281/zenodo.23422303}),
v1.1 (\path{10.5281/zenodo.23432124}), v1.2 (\path{10.5281/zenodo.23434782}), 2.0
(\path{10.5281/zenodo.23437444}) and 2.1 (\path{10.5281/zenodo.23502580}), with the concept DOI \path{10.5281/zenodo.23422302}. The commit history of the
first run is published in v1.2. Version v1.2 also contains the scorers (score.py and score\_v2.py), the analysis scripts
(analysis\_v1.py and run\_analysis.py) and the results file RESULTS.md, which gives the preregistered results not
repeated in this paper: the one-sided lower bounds of $d$ and $s$, the intervals of the class shares in S1 and S3, the
number of dropped resamples per model, the FS2 counts per age and examples of uninterpretable answers, with the numbers
in the quoted answers redacted. Version 2.0 contains the preregistration of the second run, its lock and the locked
code. The results of the second run (RESULTS\_V2.md, AVVIK\_V2.md, ERRATA\_V2.md, the run log V2\_KJORELOGG.md and the
per-answer classes ITEM\_KLASSER\_V2.csv) are in version 2.1 (\path{10.5281/zenodo.23502580}). All Zenodo versions are
published under the licence CC BY 4.0. The supplementary analyses written after the language-model review (supplement
A0--A7, not preregistered) are in version 2.1 and in the ancillary file anc/TILLEGG\_ETTER\_VURDERING.md of this arXiv
submission. Errata to the files of version 2.1 are in the ancillary file anc/ERRATA\_V2\_1.md. The uninterpretable counts of the
H3 task are in the ancillary file anc/H3\_UTOLKET\_V2.md, written from ITEM\_KLASSER\_V2.csv by
\path{scripts/h3_utolket_v2.py}. Rows of anc/TALLKILDER.md for DOIs, version numbers, publication times, the licence
and the dates of pull request 105 point at anc/OFFENTLIGE\_OPPSLAG.md, which records the public Zenodo and GitHub API
responses. The 13 rows that point at a file neither in a Zenodo version nor among the ancillary files (all of them at
ERRATA.md, the errata of the first run, which is in none of the file lists of the Zenodo versions in
anc/OFFENTLIGE\_OPPSLAG.md) can be checked under the access terms above.

\section{AI-use statement}\label{sec:ai}

Claude Code (Anthropic) wrote drafts of both preregistrations, the addendum, the run protocol, scorer v2, the reference
analyses \texttt{analysis\_v1.py} and \texttt{analysis\_v2.py}, the analysis scripts and this paper; the author read and
approved each of them. Claude Code also generated the tables and the figure, traced the numbers, orchestrated the model
runs and categorised the sampled answers in Appendix~\ref{app:explore}. The blind cross-checks of the benchmark and the
audit of the fourth lock were done by language-model sub-agents directed by the author. The review of the first version
of this paper (the ancillary file anc/FAGFELLEVURDERING.md), after which the supplementary analyses and the second
preregistration were written, was done by a language model (Claude Fable, run as a Claude Code sub-agent on 17 October
2026), not by a human reviewer. Every number in the paper was checked mechanically against the source files (the
ancillary file anc/TALLKILDER.md, 1\,549 entries, 0 mismatches); the author reviewed the text before submission.

The setting ``Help improve our AI models'' on the author's claude.ai account, which was used with Claude Code in this work,
was on when it was read on 17 October 2026 at 10:48 UTC. The settings page shows no history, so the setting during the
sessions of 16--17 October 2026 cannot be established. The setting was switched off at the author's instruction at 13:26 UTC
on 17 October 2026; it was read as still on at 13:26:19 UTC and as off at 13:26:30 UTC, and the record has no later
reading. Anthropic's privacy article on the setting
reads: ``If you decide to turn off the model training setting, we will not use any new chats and coding sessions you have
with Claude for future model training. Your data will still be included in model training that has already started and in
models that have already been trained, but we will stop using your previously stored chats and coding sessions in future
model training runs'' \citep{anthropicprivacytraining}. The item files were first committed on 16 October 2026 at 12:04:29 UTC (main items) and 12:11:04
UTC (H3 pairs of the first run), and on 17 October 2026 at 12:17:14 UTC (H3 pairs of the second run). The commits of the main
items and of the H3 pairs of the first run precede the first reading of the setting (10:48:43 UTC on 17 October 2026),
and the state of the setting before that reading is not known. The commit of the H3 pairs of the second run falls
between the readings at 10:48:43 and 13:26:19 UTC, at both of which the setting was on. For the claim that the set is closed, this means: the items are not published, but the record cannot establish
whether sessions that contained them were used in model training that started before the setting was turned off. The
record of the setting and of the commit times is in the ancillary file anc/TRENINGSINNSTILLING.md. The
model runs of the Anthropic models used the Anthropic API, whose commercial terms state that ``Anthropic may not train
models on Customer Content from Services'' \citep{anthropiccommercial}. The \mname{Gemma 4} models of the second run were run
through the unpaid quota of the Gemini API with a billing profile in Norway, an EEA country. Google's terms for the
Gemini API, as fetched before the runs, state that ``If you're in the European Economic Area, Switzerland, or the
United Kingdom, the terms under \textquotedblleft{}How Google uses Your Data\textquotedblright{} in
\textquotedblleft{}Paid Services\textquotedblright{} apply to all Services, including Google AI Studio and unpaid quota in
the Gemini API, even though they are offered free of charge'', and that ``When you use Paid Services, including, for
example, the paid quota of the Gemini API, Google doesn't use your prompts (including associated system instructions,
cached content, and files such as images, videos, or documents) or responses to improve our products''
\citep{googlegeminiterms}. For the claim that the set is closed, this means: if the runs count as made from the EEA,
Google does not use the prompts and responses of the \mname{Gemma 4} runs to improve its products under these terms. The stated cutoffs of the API
models of the first run (July 2025, January 2026 and June 2026) and of the second run (from January 2025 for the \mname{Gemma 4}
models to January 2026 for \mname{Sonnet 4.6}) precede the building of the items; the local models state no cutoff and
were run from local weight files.

\section*{Competing interests}

The author is the founder of EcoDeco AS, which offers evaluation of language-model systems, including evaluation
against the held-out items.

\bibliographystyle{plainnat}
\bibliography{refs}

\begin{thebibliography}{37}
\providecommand{\natexlab}[1]{#1}
\providecommand{\url}[1]{\texttt{#1}}
\expandafter\ifx\csname urlstyle\endcsname\relax
  \providecommand{\doi}[1]{doi: #1}\else
  \providecommand{\doi}{doi: \begingroup \urlstyle{rm}\Url}\fi

\bibitem[{Anthropic}(2026{\natexlab{a}})]{anthropiccommercial}
{Anthropic}.
\newblock Commercial terms of service, 2026{\natexlab{a}}.
\newblock URL \url{https://www.anthropic.com/legal/commercial-terms}.
\newblock Accessed 17 October 2026.

\bibitem[{Anthropic}(2026{\natexlab{b}})]{anthropicmodels}
{Anthropic}.
\newblock {Models overview}.
\newblock Claude Platform Docs, 2026{\natexlab{b}}.
\newblock URL
  \url{https://platform.claude.com/docs/en/about-claude/models/overview}.

\bibitem[{Anthropic}(2026{\natexlab{c}})]{anthropicprivacytraining}
{Anthropic}.
\newblock {How do I change my model improvement privacy settings?}
\newblock Anthropic Privacy Center, 2026{\natexlab{c}}.
\newblock URL
  \url{https://privacy.claude.com/en/articles/12109829-how-do-i-change-my-model-improvement-privacy-settings}.
\newblock Dated August 3, 2026. Accessed 17 October 2026.

\bibitem[{Anthropic}(2026{\natexlab{d}})]{haiku45page}
{Anthropic}.
\newblock {Claude Haiku 4.5}.
\newblock Claude Platform Docs, model page, 2026{\natexlab{d}}.
\newblock URL
  \url{https://platform.claude.com/docs/en/models/haiku-4-5/overview}.

\bibitem[{Anthropic}(2026{\natexlab{e}})]{opus55page}
{Anthropic}.
\newblock {Claude Opus 5.5}.
\newblock Claude Platform Docs, model page, 2026{\natexlab{e}}.
\newblock URL
  \url{https://platform.claude.com/docs/en/models/opus-5-5/overview}.

\bibitem[{Anthropic}(2026{\natexlab{f}})]{sonnet46page}
{Anthropic}.
\newblock {Claude Sonnet 4.6}.
\newblock Claude Platform Docs, model page, 2026{\natexlab{f}}.
\newblock URL
  \url{https://platform.claude.com/docs/en/models/sonnet-4-6/overview}.

\bibitem[Chen et~al.(2021)Chen, Wang, and Wang]{chen2021timeqa}
Wenhu Chen, Xinyi Wang, and William~Yang Wang.
\newblock {A Dataset for Answering Time-Sensitive Questions}.
\newblock arXiv:2108.06314, 2021.
\newblock URL \url{https://arxiv.org/abs/2108.06314}.

\bibitem[Cheng et~al.(2024)Cheng, Marone, Weller, Lawrie, Khashabi, and
  Van~Durme]{cheng2024dated}
Jeffrey Cheng, Marc Marone, Orion Weller, Dawn Lawrie, Daniel Khashabi, and
  Benjamin Van~Durme.
\newblock {Dated Data: Tracing Knowledge Cutoffs in Large Language Models}.
\newblock arXiv:2403.12958, 2024.
\newblock URL \url{https://arxiv.org/abs/2403.12958}.

\bibitem[Dahl et~al.(2024)Dahl, Magesh, Suzgun, and Ho]{dahl2024fictions}
Matthew Dahl, Varun Magesh, Mirac Suzgun, and Daniel~E. Ho.
\newblock {Large Legal Fictions: Profiling Legal Hallucinations in Large
  Language Models}.
\newblock arXiv:2401.01301, 2024.
\newblock URL \url{https://arxiv.org/abs/2401.01301}.

\bibitem[Dhingra et~al.(2021)Dhingra, Cole, Eisenschlos, Gillick, Eisenstein,
  and Cohen]{dhingra2021timeaware}
Bhuwan Dhingra, Jeremy~R. Cole, Julian~Martin Eisenschlos, Daniel Gillick,
  Jacob Eisenstein, and William~W. Cohen.
\newblock {Time-Aware Language Models as Temporal Knowledge Bases}.
\newblock arXiv:2106.15110, 2021.
\newblock URL \url{https://arxiv.org/abs/2106.15110}.

\bibitem[Field and Welsh(2007)]{field2007bootstrapping}
C.~A. Field and A.~H. Welsh.
\newblock Bootstrapping clustered data.
\newblock \emph{Journal of the Royal Statistical Society Series B: Statistical
  Methodology}, 69\penalty0 (3):\penalty0 369--390, 2007.
\newblock \doi{10.1111/j.1467-9868.2007.00593.x}.
\newblock URL \url{https://doi.org/10.1111/j.1467-9868.2007.00593.x}.

\bibitem[Gautam et~al.(2026)Gautam, Schwall, Olstad, Ruiz, Torpmann-Hagen,
  Bj{\o}rklund, Moonen, Pettersen, and Riegler]{gautam2026benchmarkless}
Sushant Gautam, Finn Schwall, Annika~Willoch Olstad, Fernando~Vallecillos Ruiz,
  Birk Torpmann-Hagen, Sunniva Maria~Stordal Bj{\o}rklund, Leon Moonen, Klas
  Pettersen, and Michael~A. Riegler.
\newblock {When No Benchmark Exists: Validating Comparative LLM Safety Scoring
  Without Ground-Truth Labels}.
\newblock arXiv:2605.06652, 2026.
\newblock URL \url{https://arxiv.org/abs/2605.06652}.

\bibitem[{Google}(2026)]{googlegeminiterms}
{Google}.
\newblock {Gemini API Additional Terms of Service}.
\newblock Google AI for Developers, fetched 17 October, 2026.
\newblock URL \url{https://ai.google.dev/gemini-api/terms}.

\bibitem[Guha et~al.(2023)Guha, Nyarko, Ho, Ré, Chilton, Narayana,
  Chohlas-Wood, Peters, Waldon, Rockmore, Zambrano, Talisman, Hoque, Surani,
  Fagan, Sarfaty, Dickinson, Porat, Hegland, Wu, Nudell, Niklaus, Nay, Choi,
  Tobia, Hagan, Ma, Livermore, Rasumov-Rahe, Holzenberger, Kolt, Henderson,
  Rehaag, Goel, Gao, Williams, Gandhi, Zur, Iyer, and Li]{guha2023legalbench}
Neel Guha, Julian Nyarko, Daniel~E. Ho, Christopher Ré, Adam Chilton, Aditya
  Narayana, Alex Chohlas-Wood, Austin Peters, Brandon Waldon, Daniel~N.
  Rockmore, Diego Zambrano, Dmitry Talisman, Enam Hoque, Faiz Surani, Frank
  Fagan, Galit Sarfaty, Gregory~M. Dickinson, Haggai Porat, Jason Hegland,
  Jessica Wu, Joe Nudell, Joel Niklaus, John Nay, Jonathan~H. Choi, Kevin
  Tobia, Margaret Hagan, Megan Ma, Michael Livermore, Nikon Rasumov-Rahe, Nils
  Holzenberger, Noam Kolt, Peter Henderson, Sean Rehaag, Sharad Goel, Shang
  Gao, Spencer Williams, Sunny Gandhi, Tom Zur, Varun Iyer, and Zehua Li.
\newblock {LegalBench: A Collaboratively Built Benchmark for Measuring Legal
  Reasoning in Large Language Models}.
\newblock arXiv:2308.11462, 2023.
\newblock URL \url{https://arxiv.org/abs/2308.11462}.

\bibitem[Hofman et~al.(2023)Hofman, Chatzimparmpas, Sharma, Watts, and
  Hullman]{hofman2023prereg}
Jake~M. Hofman, Angelos Chatzimparmpas, Amit Sharma, Duncan~J. Watts, and
  Jessica Hullman.
\newblock {Pre-registration for Predictive Modeling}.
\newblock arXiv:2311.18807, 2023.
\newblock URL \url{https://arxiv.org/abs/2311.18807}.

\bibitem[Holm(1979)]{holm1979}
Sture Holm.
\newblock A simple sequentially rejective multiple test procedure.
\newblock \emph{Scandinavian Journal of Statistics}, 6\penalty0 (2):\penalty0
  65--70, 1979.
\newblock URL \url{https://www.jstor.org/stable/4615733}.

\bibitem[Jang et~al.(2022)Jang, Ye, Lee, Yang, Shin, Han, Kim, and
  Seo]{jang2022temporalwiki}
Joel Jang, Seonghyeon Ye, Changho Lee, Sohee Yang, Joongbo Shin, Janghoon Han,
  Gyeonghun Kim, and Minjoon Seo.
\newblock {TemporalWiki: A Lifelong Benchmark for Training and Evaluating
  Ever-Evolving Language Models}.
\newblock arXiv:2204.14211, 2022.
\newblock URL \url{https://arxiv.org/abs/2204.14211}.

\bibitem[Kadavath et~al.(2022)Kadavath, Conerly, Askell, Henighan, Drain,
  Perez, Schiefer, Hatfield-Dodds, DasSarma, Tran-Johnson, Johnston, El-Showk,
  Jones, Elhage, Hume, Chen, Bai, Bowman, Fort, Ganguli, Hernandez, Jacobson,
  Kernion, Kravec, Lovitt, Ndousse, Olsson, Ringer, Amodei, Brown, Clark,
  Joseph, Mann, McCandlish, Olah, and Kaplan]{kadavath2022know}
Saurav Kadavath, Tom Conerly, Amanda Askell, Tom Henighan, Dawn Drain, Ethan
  Perez, Nicholas Schiefer, Zac Hatfield-Dodds, Nova DasSarma, Eli
  Tran-Johnson, Scott Johnston, Sheer El-Showk, Andy Jones, Nelson Elhage,
  Tristan Hume, Anna Chen, Yuntao Bai, Sam Bowman, Stanislav Fort, Deep
  Ganguli, Danny Hernandez, Josh Jacobson, Jackson Kernion, Shauna Kravec,
  Liane Lovitt, Kamal Ndousse, Catherine Olsson, Sam Ringer, Dario Amodei, Tom
  Brown, Jack Clark, Nicholas Joseph, Ben Mann, Sam McCandlish, Chris Olah, and
  Jared Kaplan.
\newblock {Language Models (Mostly) Know What They Know}.
\newblock arXiv:2207.05221, 2022.
\newblock URL \url{https://arxiv.org/abs/2207.05221}.

\bibitem[Kasai et~al.(2022)Kasai, Sakaguchi, Takahashi, Bras, Asai, Yu, Radev,
  Smith, Choi, and Inui]{kasai2022realtimeqa}
Jungo Kasai, Keisuke Sakaguchi, Yoichi Takahashi, Ronan~Le Bras, Akari Asai,
  Xinyan Yu, Dragomir Radev, Noah~A. Smith, Yejin Choi, and Kentaro Inui.
\newblock {RealTime QA: What's the Answer Right Now?}
\newblock arXiv:2207.13332, 2022.
\newblock URL \url{https://arxiv.org/abs/2207.13332}.

\bibitem[{Language Technology Group (ltg)}(2024)]{normistralcard}
{Language Technology Group (ltg)}.
\newblock {ltg/normistral-7b-warm-instruct}.
\newblock Hugging Face model card, Apache-2.0, 2024.
\newblock URL \url{https://huggingface.co/ltg/normistral-7b-warm-instruct}.

\bibitem[Lazaridou et~al.(2021)Lazaridou, Kuncoro, Gribovskaya, Agrawal, Liska,
  Terzi, Gimenez, d'Autume, Kocisky, Ruder, Yogatama, Cao, Young, and
  Blunsom]{lazaridou2021mindthegap}
Angeliki Lazaridou, Adhiguna Kuncoro, Elena Gribovskaya, Devang Agrawal, Adam
  Liska, Tayfun Terzi, Mai Gimenez, Cyprien de~Masson d'Autume, Tomas Kocisky,
  Sebastian Ruder, Dani Yogatama, Kris Cao, Susannah Young, and Phil Blunsom.
\newblock {Mind the Gap: Assessing Temporal Generalization in Neural Language
  Models}.
\newblock arXiv:2102.01951, 2021.
\newblock URL \url{https://arxiv.org/abs/2102.01951}.

\bibitem[Liu et~al.(2023)Liu, Zhang, Farup, Lauvrak, Ingvaldsen, Eide, Gulla,
  and Yang]{liu2023nlebench}
Peng Liu, Lemei Zhang, Terje Farup, Even~W. Lauvrak, Jon~Espen Ingvaldsen,
  Simen Eide, Jon~Atle Gulla, and Zhirong Yang.
\newblock {NLEBench+NorGLM: A Comprehensive Empirical Analysis and Benchmark
  Dataset for Generative Language Models in Norwegian}.
\newblock arXiv:2312.01314, 2023.
\newblock URL \url{https://arxiv.org/abs/2312.01314}.

\bibitem[Liška et~al.(2022)Liška, Kočiský, Gribovskaya, Terzi, Sezener,
  Agrawal, d'Autume, Scholtes, Zaheer, Young, Gilsenan-McMahon, Austin,
  Blunsom, and Lazaridou]{liska2022streamingqa}
Adam Liška, Tomáš Kočiský, Elena Gribovskaya, Tayfun Terzi, Eren Sezener,
  Devang Agrawal, Cyprien de~Masson d'Autume, Tim Scholtes, Manzil Zaheer,
  Susannah Young, Ellen Gilsenan-McMahon, Sophia Austin, Phil Blunsom, and
  Angeliki Lazaridou.
\newblock {StreamingQA: A Benchmark for Adaptation to New Knowledge over Time
  in Question Answering Models}.
\newblock arXiv:2205.11388, 2022.
\newblock URL \url{https://arxiv.org/abs/2205.11388}.

\bibitem[Magesh et~al.(2024)Magesh, Surani, Dahl, Suzgun, Manning, and
  Ho]{magesh2024hallucination}
Varun Magesh, Faiz Surani, Matthew Dahl, Mirac Suzgun, Christopher~D. Manning,
  and Daniel~E. Ho.
\newblock {Hallucination-Free? Assessing the Reliability of Leading AI Legal
  Research Tools}.
\newblock arXiv:2405.20362, 2024.
\newblock URL \url{https://arxiv.org/abs/2405.20362}.

\bibitem[McNemar(1947)]{mcnemar1947}
Quinn McNemar.
\newblock Note on the sampling error of the difference between correlated
  proportions or percentages.
\newblock \emph{Psychometrika}, 12\penalty0 (2):\penalty0 153--157, 1947.
\newblock \doi{10.1007/BF02295996}.
\newblock URL \url{https://doi.org/10.1007/BF02295996}.

\bibitem[Mikhailov et~al.(2025)Mikhailov, Enstad, Samuel, Farsethås, Kutuzov,
  Velldal, and Øvrelid]{mikhailov2025noreval}
Vladislav Mikhailov, Tita Enstad, David Samuel, Hans~Christian Farsethås,
  Andrey Kutuzov, Erik Velldal, and Lilja Øvrelid.
\newblock {NorEval: A Norwegian Language Understanding and Generation
  Evaluation Benchmark}.
\newblock arXiv:2504.07749, 2025.
\newblock URL \url{https://arxiv.org/abs/2504.07749}.

\bibitem[{NbAiLab}(2024)]{nbllamacard}
{NbAiLab}.
\newblock {NbAiLab/nb-llama-3.1-8B-Instruct-Q4\_K\_M-GGUF}.
\newblock Hugging Face model card, base model NB-Llama-3.1-8B-Instruct, Llama
  3.1 license, 2024.
\newblock URL
  \url{https://huggingface.co/NbAiLab/nb-llama-3.1-8B-Instruct-Q4_K_M-GGUF}.

\bibitem[Nicolaysen(2026)]{simpleaudit105}
Eirik~Botten Nicolaysen.
\newblock {Add a deterministic fact\_check judge over metadata.facts}.
\newblock SimulaMet/SimpleAudit, GitHub pull request 105 (GitHub user
  avalyset), opened 9 October, merged into the dev branch 10 October, 2026.
\newblock URL \url{https://github.com/SimulaMet/SimpleAudit/pull/105}.

\bibitem[Nosek et~al.(2018)Nosek, Ebersole, DeHaven, and
  Mellor]{nosek2018prereg}
Brian~A. Nosek, Charles~R. Ebersole, Alexander~C. DeHaven, and David~T. Mellor.
\newblock The preregistration revolution.
\newblock \emph{Proceedings of the National Academy of Sciences}, 115\penalty0
  (11):\penalty0 2600--2606, 2018.
\newblock \doi{10.1073/pnas.1708274114}.
\newblock URL \url{https://doi.org/10.1073/pnas.1708274114}.

\bibitem[Samuel et~al.(2023)Samuel, Kutuzov, Touileb, Velldal, {\O}vrelid,
  R{\o}nningstad, Sigdel, and Palatkina]{samuel2023norbench}
David Samuel, Andrey Kutuzov, Samia Touileb, Erik Velldal, Lilja {\O}vrelid,
  Egil R{\o}nningstad, Elina Sigdel, and Anna Palatkina.
\newblock {N}or{B}ench {--} a benchmark for {N}orwegian language models.
\newblock In \emph{Proceedings of the 24th {N}ordic Conference on Computational
  Linguistics ({N}o{D}a{L}i{D}a)}, pages 618--633, T{\'o}rshavn, Faroe Islands,
  May 2023. University of Tartu Library.
\newblock URL \url{https://aclanthology.org/2023.nodalida-1.61/}.

\bibitem[{SimulaMet}(2026)]{simpleauditrepo}
{SimulaMet}.
\newblock {SimpleAudit}.
\newblock GitHub repository SimulaMet/SimpleAudit, README read 20 October,
  2026.
\newblock URL \url{https://github.com/SimulaMet/SimpleAudit}.

\bibitem[Smart(2023)]{nielsen2023scandeval}
Dan~Saattrup Smart.
\newblock {S}cand{E}val: A benchmark for {S}candinavian natural language
  processing.
\newblock In \emph{Proceedings of the 24th {N}ordic Conference on Computational
  Linguistics ({N}o{D}a{L}i{D}a)}, pages 185--201, T{\'o}rshavn, Faroe Islands,
  May 2023. University of Tartu Library.
\newblock URL \url{https://aclanthology.org/2023.nodalida-1.20/}.

\bibitem[{UKGovernmentBEIS}(2026)]{inspect}
{UKGovernmentBEIS}.
\newblock {Inspect: A framework for large language model evaluations}.
\newblock GitHub repository UKGovernmentBEIS/inspect\_ai, MIT license, 2026.
\newblock URL \url{https://github.com/UKGovernmentBEIS/inspect_ai}.
\newblock Version 0.3.245 used.

\bibitem[van Miltenburg et~al.(2021)van Miltenburg, van~der Lee, and
  Krahmer]{vanmiltenburg2021prereg}
Emiel van Miltenburg, Chris van~der Lee, and Emiel Krahmer.
\newblock Preregistering {NLP} research.
\newblock In \emph{Proceedings of the 2021 Conference of the {N}orth {A}merican
  Chapter of the {A}ssociation for {C}omputational {L}inguistics: Human
  Language Technologies}, pages 613--623, Online, June 2021. Association for
  Computational Linguistics.
\newblock \doi{10.18653/v1/2021.naacl-main.51}.
\newblock URL \url{https://aclanthology.org/2021.naacl-main.51/}.

\bibitem[Vu et~al.(2023)Vu, Iyyer, Wang, Constant, Wei, Wei, Tar, Sung, Zhou,
  Le, and Luong]{vu2023freshllms}
Tu~Vu, Mohit Iyyer, Xuezhi Wang, Noah Constant, Jerry Wei, Jason Wei, Chris
  Tar, Yun-Hsuan Sung, Denny Zhou, Quoc Le, and Thang Luong.
\newblock {FreshLLMs: Refreshing Large Language Models with Search Engine
  Augmentation}.
\newblock arXiv:2310.03214, 2023.
\newblock URL \url{https://arxiv.org/abs/2310.03214}.

\bibitem[Yin et~al.(2023)Yin, Sun, Guo, Wu, Qiu, and Huang]{yin2023selfaware}
Zhangyue Yin, Qiushi Sun, Qipeng Guo, Jiawen Wu, Xipeng Qiu, and Xuanjing
  Huang.
\newblock {Do Large Language Models Know What They Don't Know?}
\newblock arXiv:2305.18153, 2023.
\newblock URL \url{https://arxiv.org/abs/2305.18153}.

\bibitem[Zhang and Choi(2021)]{zhang2021situatedqa}
Michael J.~Q. Zhang and Eunsol Choi.
\newblock {SituatedQA: Incorporating Extra-Linguistic Contexts into QA}.
\newblock arXiv:2109.06157, 2021.
\newblock URL \url{https://arxiv.org/abs/2109.06157}.

\end{thebibliography}

\appendix

\section{Exploratory revision of the uninterpretable class (not preregistered)}\label{app:explore}

This analysis was chosen after the results were known. It is not preregistered, and it changes neither the primary
scoring nor any locked file. For \mname{Haiku 4.5}, \mname{Sonnet 4.6} and \mname{NB-Llama 8B}, 30 uninterpretable
answers in the primary set were drawn at random per model and read in full by one reader (Claude Code), without the
keys, and placed in categories (a)--(e) (Table~\ref{tab:explore}). For \mname{NB-Llama 8B}, 9 of 30 answers (30\%,
Clopper--Pearson 95\% CI 14.7--49.4\%) gave one value in a format the scorer does not read. This met the trigger of
20\% set for the revision, and the primary set was re-scored for all five models with a tolerant parser that accepts
only the three formats found (Table~\ref{tab:tolerant}). The parser rewrote 113 of the 604 uninterpretable answers of
\mname{NB-Llama 8B} and 11 answers of \mname{NorMistral 7B}, and none for the API models. The formats were taken from
the same answers, so this is not an independent test, and none of these numbers replaces Table~\ref{tab:primary}.

\begin{table}[t]
\centering
\footnotesize
\caption{Exploratory, not preregistered: categories of a random sample of uninterpretable answers. (a) refusal; (b) no value; (c) one value in a format the scorer does not read; (d) several values or ambiguous; (e) other.}
\label{tab:explore}
\begin{tabular}{llllllll}
\toprule
Model & Uninterpretable & (a) & (b) & (c) & (d) & (e) & Sample \\
\midrule
\mname{Haiku 4.5} & 827 & 30 & 0 & 0 & 0 & 0 & 30 \\
\mname{Sonnet 4.6} & 109 & 30 & 0 & 0 & 0 & 0 & 30 \\
\mname{NB-Llama 8B} & 604 & 13 & 0 & 9 & 4 & 4 & 30 \\
\bottomrule
\end{tabular}
\end{table}

\begin{table}[t]
\centering
\footnotesize
\caption{Exploratory, not preregistered: primary set re-scored after a tolerant parser rewrote answers in three formats found in category (c). Shares (\%) and number of rewritten answers.}
\label{tab:tolerant}
\begin{tabular}{llllll}
\toprule
Model & Correct & Outdated & Wrong & Uninterpretable & Rewritten \\
\midrule
\mname{Haiku 4.5} & 0.5 & 0.1 & 5.5 & 94.0 & 0 \\
\mname{Sonnet 4.6} & 40.2 & 6.5 & 40.9 & 12.4 & 0 \\
\mname{Opus 5.5} & 46.7 & 6.5 & 46.4 & 0.5 & 0 \\
\mname{NB-Llama 8B} & 11.5 & 1.7 & 31.0 & 55.8 & 113 \\
\mname{NorMistral 7B} & 0.5 & 0.2 & 0.8 & 98.5 & 11 \\
\bottomrule
\end{tabular}
\end{table}

\section{Preregistered reporting: exclusion groups and other runs}\label{app:prereg}

The addendum requires each exclusion group to be reported with all four classes per model (Table~\ref{tab:groups}),
and every run to be reported (Table~\ref{tab:otherruns}).

\begin{table}[t]
\centering
\footnotesize
\caption{Excluded items per exclusion group (counts, scorer v2), as preregistered in the addendum. X1 wrong key for the question date; X2 question date after the corpus retrieval; X3 ambiguous value in the provision; X4 amendments of the annual tax decision.}
\label{tab:groups}
\begin{tabular}{lllllll}
\toprule
Model & Group & $n$ & Correct & Outdated & Wrong & Uninterpretable \\
\midrule
\mname{Haiku 4.5} & X1 & 5 & 0 & 0 & 0 & 5 \\
\mname{Haiku 4.5} & X2 & 23 & 0 & 0 & 0 & 23 \\
\mname{Haiku 4.5} & X3 & 6 & 0 & 0 & 0 & 6 \\
\mname{Haiku 4.5} & X4 & 2 & 0 & 0 & 0 & 2 \\
\mname{Sonnet 4.6} & X1 & 5 & 1 & 1 & 3 & 0 \\
\mname{Sonnet 4.6} & X2 & 23 & 0 & 0 & 0 & 23 \\
\mname{Sonnet 4.6} & X3 & 6 & 3 & 0 & 3 & 0 \\
\mname{Sonnet 4.6} & X4 & 2 & 0 & 0 & 2 & 0 \\
\mname{Opus 5.5} & X1 & 5 & 0 & 2 & 3 & 0 \\
\mname{Opus 5.5} & X2 & 23 & 0 & 0 & 8 & 15 \\
\mname{Opus 5.5} & X3 & 6 & 1 & 0 & 3 & 2 \\
\mname{Opus 5.5} & X4 & 2 & 1 & 0 & 1 & 0 \\
\mname{NB-Llama 8B} & X1 & 5 & 0 & 0 & 2 & 3 \\
\mname{NB-Llama 8B} & X2 & 23 & 0 & 0 & 0 & 23 \\
\mname{NB-Llama 8B} & X3 & 6 & 0 & 0 & 2 & 4 \\
\mname{NB-Llama 8B} & X4 & 2 & 0 & 0 & 0 & 2 \\
\mname{NorMistral 7B} & X1 & 5 & 0 & 0 & 0 & 5 \\
\mname{NorMistral 7B} & X2 & 23 & 0 & 0 & 0 & 23 \\
\mname{NorMistral 7B} & X3 & 6 & 0 & 0 & 0 & 6 \\
\mname{NorMistral 7B} & X4 & 2 & 0 & 0 & 0 & 2 \\
\bottomrule
\end{tabular}
\end{table}

\begin{table*}[t]
\centering
\tiny\setlength{\tabcolsep}{3pt}
\caption{Every run that is not a primary log, all on the main task: start (UTC) from the log header, status, samples in the log, samples in error, and classes under scorer v2 (correct / outdated / wrong / uninterpretable) of the samples present. None of them counts in the results (supplement A1).}
\label{tab:otherruns}
\begin{tabular}{lllllll}
\toprule
Start (UTC) & Model & Status & Samples & Errors & Classes & Reason \\
\midrule
2026-10-16 16:30:54 & {\tiny\path{anthropic/claude-haiku-4-5-20251001}} & success & 20 & 20 & 0 / 0 / 0 / 20 & pilot, A1 \\
2026-10-16 16:31:39 & {\tiny\path{anthropic/claude-haiku-4-5-20251001}} & success & 20 & 0 & 0 / 0 / 1 / 19 & pilot \\
2026-10-16 16:32:18 & {\tiny\path{anthropic/claude-opus-5-5}} & success & 20 & 0 & 10 / 1 / 9 / 0 & pilot \\
2026-10-16 16:32:01 & {\tiny\path{anthropic/claude-sonnet-4-6}} & success & 20 & 0 & 9 / 1 / 8 / 2 & pilot \\
2026-10-16 16:41:54 & {\tiny\path{ollama/hf.co/ltg/normistral-7b-warm-instruct:Q4_K_M}} & success & 20 & 0 & 0 / 0 / 0 / 20 & pilot \\
2026-10-16 17:52:39 & {\tiny\path{ollama/nb-llama-3.1-8b-instruct-cardtmpl:latest}} & success & 20 & 0 & 1 / 1 / 4 / 14 & pilot \\
2026-10-16 16:33:05 & {\tiny\path{ollama/nb-llama-3.1-8b:latest}} & success & 20 & 0 & 1 / 0 / 4 / 15 & pilot \\
2026-10-16 17:49:17 & {\tiny\path{ollama/normistral-7b-warm-instruct-cardtmpl:latest}} & success & 20 & 0 & 0 / 0 / 0 / 20 & pilot \\
2026-10-16 17:51:32 & {\tiny\path{ollama/normistral-7b-warm-instruct-cardtmpl:latest}} & cancelled & 5 & 1 & 0 / 0 / 0 / 5 & stopped, A4/A5 \\
2026-10-16 16:45:40 & {\tiny\path{ollama/nb-llama-3.1-8b:latest}} & cancelled & 109 & 1 & 6 / 3 / 19 / 81 & discarded, A3 \\
2026-10-17 01:12:43 & {\tiny\path{ollama/normistral-7b-warm-instruct-cardtmpl:latest}} & cancelled & 92 & 1 & 0 / 0 / 0 / 92 & first attempt, A6 \\
\bottomrule
\end{tabular}
\end{table*}

\section{Analyses after the language-model review (not preregistered)}\label{app:review}

These tables were computed after the language-model review (Section~\ref{sec:ai}) from the existing answers and metadata (supplement A3 and A4). They are
exploratory and replace no preregistered result. The appendix has Tables~\ref{tab:interp}, \ref{tab:h1groups} and
\ref{tab:h3cond}.

\begin{table}[t]
\centering
\footnotesize
\caption{Exploratory, after review: classes among interpretable answers (correct, outdated or wrong), primary set (supplement A3). Shares in \%.}
\label{tab:interp}
\begin{tabular}{lllll}
\toprule
Model & Interpretable / all & Correct & Outdated & Wrong \\
\midrule
\mname{Haiku 4.5} & 53 / 880 & 7.5 & 1.9 & 90.6 \\
\mname{Sonnet 4.6} & 771 / 880 & 45.9 & 7.4 & 46.7 \\
\mname{Opus 5.5} & 876 / 880 & 46.9 & 6.5 & 46.6 \\
\mname{NB-Llama 8B} & 276 / 880 & 27.2 & 4.0 & 68.8 \\
\mname{NorMistral 7B} & 2 / 880 & 0.0 & 0.0 & 100.0 \\
\bottomrule
\end{tabular}
\end{table}

\begin{table*}[t]
\centering
\footnotesize\setlength{\tabcolsep}{4pt}
\caption{Exploratory, after review: uninterpretable answers in the H1 set after and before the cutoff (count and \% of all answers in the group), and the classes after the cutoff (correct / outdated / wrong / uninterpretable); supplement A3.}
\label{tab:h1groups}
\begin{tabular}{llllll}
\toprule
Model & Cutoff & $n$ after / before & Uninterpretable after & Uninterpretable before & Classes after \\
\midrule
\mname{Haiku 4.5} & 2025-07-31 & 56 / 511 & 56 (100.0) & 475 (93.0) & 0 / 0 / 0 / 56 \\
\mname{Sonnet 4.6} & 2026-01-31 & 9 / 558 & 8 (88.9) & 74 (13.3) & 0 / 0 / 1 / 8 \\
\mname{Opus 5.5} & 2026-06-30 & 1 / 566 & 0 (0.0) & 4 (0.7) & 1 / 0 / 0 / 0 \\
\bottomrule
\end{tabular}
\end{table*}

\begin{table}[t]
\centering
\footnotesize
\caption{Exploratory, after review: H3 difference (Nynorsk $-$ Bokmål, pp) with an exact interval from $b$ and $c$ alone (Clopper--Pearson for $c/(b+c)$, converted to $(c-b)/n$ given $b+c$; it ignores clustering and is narrower), next to the preregistered bootstrap interval; supplement A4.}
\label{tab:h3cond}
\begin{tabular}{llllll}
\toprule
Model & $b$ & $c$ & Difference & Interval from $b$, $c$ & Bootstrap interval \\
\midrule
\mname{Haiku 4.5} & 1 & 0 & $-$0.5 & [$-$0.5, 0.5] & [$-$1.6, 0.0] \\
\mname{Sonnet 4.6} & 9 & 3 & $-$3.2 & [$-$5.7, 0.9] & [$-$8.6, 1.1] \\
\mname{Opus 5.5} & 3 & 8 & 2.6 & [$-$1.3, 5.1] & [$-$1.0, 6.6] \\
\mname{NB-Llama 8B} & 6 & 1 & $-$2.6 & [$-$3.7, 0.6] & [$-$6.7, 0.5] \\
\mname{NorMistral 7B} & 0 & 0 & 0.0 & not defined & [0.0, 0.0] \\
\bottomrule
\end{tabular}
\end{table}

\section{Further tables}

This appendix has the class shares of the sensitivity analyses of the first run (Table~\ref{tab:sensclasses}), the share
correct by value type (Table~\ref{tab:types}) and the split of wrong answers into a later value and a value not in the
history (Table~\ref{tab:wrong}).

\begin{table}[t]
\centering
\footnotesize
\caption{Sensitivity analyses: share of all answers (\%) per class. Percentages are rounded to one decimal, half to even.}
\label{tab:sensclasses}
\begin{tabular}{lllll}
\toprule
Model & Correct & Outdated & Wrong & Uninterpretable \\
\midrule
\multicolumn{5}{l}{\textit{S1 (all items)}} \\
\mname{Haiku 4.5} & 0.4 & 0.1 & 5.2 & 94.2 \\
\mname{Sonnet 4.6} & 39.1 & 6.3 & 40.2 & 14.4 \\
\mname{Opus 5.5} & 45.1 & 6.4 & 46.2 & 2.3 \\
\mname{NB-Llama 8B} & 8.2 & 1.2 & 21.2 & 69.4 \\
\mname{NorMistral 7B} & 0.0 & 0.0 & 0.2 & 99.8 \\
\multicolumn{5}{l}{\textit{S2 (scorer v1)}} \\
\mname{Haiku 4.5} & 0.5 & 0.7 & 0.1 & 98.8 \\
\mname{Sonnet 4.6} & 40.2 & 11.1 & 36.2 & 12.4 \\
\mname{Opus 5.5} & 46.7 & 11.9 & 40.1 & 1.2 \\
\mname{NB-Llama 8B} & 7.6 & 2.2 & 14.1 & 76.1 \\
\mname{NorMistral 7B} & 0.0 & 0.0 & 0.1 & 99.9 \\
\multicolumn{5}{l}{\textit{S3 (without flagged)}} \\
\mname{Haiku 4.5} & 0.5 & 0.1 & 5.5 & 93.9 \\
\mname{Sonnet 4.6} & 40.4 & 6.3 & 41.0 & 12.3 \\
\mname{Opus 5.5} & 46.5 & 6.5 & 46.5 & 0.5 \\
\mname{NB-Llama 8B} & 8.6 & 1.3 & 21.5 & 68.6 \\
\mname{NorMistral 7B} & 0.0 & 0.0 & 0.2 & 99.8 \\
\bottomrule
\end{tabular}
\end{table}

\begin{table}[t]
\centering
\footnotesize
\caption{Share correct (\%) by value type, primary set (counts and shares only, no intervals or tests). Percentages are rounded to one decimal, half to even.}
\label{tab:types}
\begin{tabular}{llll}
\toprule
Model & NOK amount ($n=553$) & Percentage ($n=319$) & Age ($n=8$) \\
\midrule
\mname{Haiku 4.5} & 0.2 & 0.6 & 12.5 \\
\mname{Sonnet 4.6} & 27.5 & 63.3 & 0.0 \\
\mname{Opus 5.5} & 32.9 & 70.8 & 37.5 \\
\mname{NB-Llama 8B} & 3.1 & 17.9 & 12.5 \\
\mname{NorMistral 7B} & 0.0 & 0.0 & 0.0 \\
\bottomrule
\end{tabular}
\end{table}

\begin{table}[t]
\centering
\footnotesize
\caption{Wrong answers split into a later value and a value not in the item's history (counts, primary set).}
\label{tab:wrong}
\begin{tabular}{llll}
\toprule
Model & Wrong & Later value & Not in history \\
\midrule
\mname{Haiku 4.5} & 48 & 5 & 43 \\
\mname{Sonnet 4.6} & 360 & 41 & 319 \\
\mname{Opus 5.5} & 408 & 49 & 359 \\
\mname{NB-Llama 8B} & 190 & 10 & 180 \\
\mname{NorMistral 7B} & 2 & 0 & 2 \\
\bottomrule
\end{tabular}
\end{table}

\end{document}